\documentclass{article}

\pdfoutput=1
\usepackage[preprint]{neurips_data_2024}

\usepackage[utf8]{inputenc} 
\usepackage[T1]{fontenc}    
\usepackage{hyperref}       
\usepackage{url}            
\usepackage{booktabs}       
\usepackage{amsfonts}       
\usepackage{nicefrac}       
\usepackage{microtype}      
\usepackage{xcolor}         
\usepackage{makecell}
\usepackage{multirow}
\usepackage{amsmath} 
\usepackage{ulem}
\usepackage{algorithm2e}
\newcommand{\VarSty}[1]{\textbf{\ttfamily\color{blue!80!black}#1}\unskip}

\newcommand{\var}{\texttt}

\SetKwProg{Function}{def}{:}{}
\SetKwProg{For}{for}{:}{}
\SetKwProg{If}{if}{:}{}
\usepackage{tcolorbox}
\usepackage{graphicx} 
\usepackage{colortbl}
\usepackage{pifont}
\usepackage{tabularx}
\usepackage{enumitem}

\title{SkySenseGPT: A Fine-Grained Instruction Tuning Dataset and Model for Remote Sensing Vision-Language Understanding}

\author{
  Junwei Luo  \\
  Wuhan University\\
  \texttt{luojunwei}\\\texttt{@whu.edu.cn}\hspace{-0.1em}\vspace{-0.6em}\\
  \And
  Zhen Pang \\
  Wuhan University \\
  \texttt{pangzhen}\\\texttt{@whu.edu.cn}\hspace{-0.2em}\vspace{-0.6em}\\
  \And
  Yongjun Zhang \\
  Wuhan University \\
  \texttt{zhangyj}\\\texttt{@whu.edu.cn}\vspace{-0.6em}\\
  \And
  Tingzhu Wang \\
  Wuhan University \\
  \texttt{tingzhu.wang}\\\texttt{@whu.edu.cn}\vspace{-0.6em}\\
  \And
  Linlin Wang \\
  Wuhan University \\
  \texttt{wangll}\\\texttt{@whu.edu.cn}\vspace{-0.6em}\\
  \And
  Bo Dang \\
  Wuhan University \\
  \texttt{bodang}\\\texttt{@whu.edu.cn}\vspace{-0.6em}\\
  \And
  Jiangwei Lao \\
  Ant Group \\
  \texttt{wenshuo.ljw}\\\texttt{@antgroup.com}\vspace{-0.6em}\\
  \And
  Jian Wang\\
  Ant Group \\
  \texttt{bobblair.wj}\\\texttt{@antgroup.com}\vspace{-0.6em}\\
  \AND
  Jingdong Chen \\
  Ant Group \\
  \texttt{jingdongchen.cjd}\\\texttt{@antgroup.com}\vspace{-0.6em}\\
  \And
  Yihua Tan \\
  Huazhong University of\\Science and Technology \\
  \texttt{yhtan@hust.edu.cn}  \\
  \And
  Yansheng Li\thanks{Corresponding author.} \\
  Wuhan University \\
  \texttt{yansheng.li}\\\texttt{@whu.edu.cn}\\}

\begin{document}

\maketitle

\begin{abstract}
Remote Sensing Large Multi-Modal Models (RSLMMs) are developing rapidly and showcase significant capabilities in remote sensing imagery (RSI) comprehension. However, due to the limitations of existing datasets, RSLMMs have shortcomings in understanding the rich semantic relations among objects in complex remote sensing scenes. To unlock RSLMMs' complex comprehension ability, we propose a large-scale instruction tuning dataset FIT-RS, containing 1,800,851 instruction samples. FIT-RS covers common interpretation tasks and innovatively introduces several complex comprehension tasks of escalating difficulty, ranging from relation reasoning to image-level scene graph generation. Based on FIT-RS, we build the FIT-RSFG benchmark. Furthermore, we establish a new benchmark to evaluate the fine-grained relation comprehension capabilities of LMMs, named FIT-RSRC. Based on combined instruction data, we propose SkySenseGPT, which achieves outstanding performance on both public datasets and FIT-RSFG, surpassing existing RSLMMs. We hope the FIT-RS dataset can enhance the relation comprehension capability of RSLMMs and provide a large-scale fine-grained data source for the remote sensing community. The dataset will be available at 
\url{https://github.com/Luo-Z13/SkySenseGPT}.

\end{abstract}

\section{Introduction}

Benefiting from the recent rapid advancements and evolution of Large Language Models (LLMs)~\cite{achiam2023gpt,touvron2023llama}, the capabilities of Large Multi-Modal Models (LMMs)~\cite{liu2024visual,wang2023all,zhu2023minigpt4,wang2024visionllm,liu2023improved,bai2023qwen,chen2023internvl,reid2024gemini} have significantly improved, yielding notable progress. Based on various large-scale visual-language datasets~\cite{kazemzadeh2014referitgame,krishna2017visual,hudson2019gqa,mani2020point,yang2022psg,liu2023visual}, existing general LMMs have demonstrated impressive capabilities in multiple tasks, including handling intricate tasks like complex reasoning~\cite{liu2023improved}, optical character recognition (OCR)~\cite{ye2023ureader,liu2024textmonkey}, scene graph generation (SGG)~\cite{wang2024all}, and so on. Moreover, LMMs have been extended to various other fields like medical assistance~\cite{xiao2024comprehensive} and autonomous driving~\cite{wang2023drivemlm,cui2024survey}.

\begin{table}[!t]
\caption{Comparison with existing remote sensing instruction tuning datasets. ``From Scratch'' means collecting \textbf{all images} from scratch instead of combining public datasets. ``Fine-Grained'' means whether the relationships are fine-grained, and explicitly used in dataset and evaluation.}
\label{tab:sample_compare}
\small
\renewcommand{\arraystretch}{1.3}
\centering
\begin{tabularx}{\textwidth}{lcccX}
\toprule
\textbf{Dataset} & \textbf{\thead{From\\Scratch}} & \textbf{\thead{Fine-\\Grained}} & \textbf{\thead{Total \\Sample}} & \textbf{Tasks}   \\ \hline
RSICap~\cite{hu2023rsgpt} & $\times$ & $\times$ & 2.6k  & Image Caption     \\
\multirow{2}{*}{H$^2$RSVLM-SFT~\cite{pang2024h2rsvlm}} & \multirow{2}{*}{$\times$} & \multirow{2}{*}{$\times$} & \multirow{2}{*}{180k} & Image Caption, VQA, Multi-Label Scene Classification, Visual Grounding, other basic tasks \\
\multirow{4}{*}{GeoChat~\cite{kuckreja2023geochat}} & \multirow{4}{*}{$\times$} & \multirow{4}{*}{$\times$} & \multirow{4}{*}{318k} & Image Caption, Region Caption, VQA, Scene Classification, Visual Grounding, Object Detection, Grounding Description, Multi-Task Conversation   \\
\multirow{3}{*}{MMRS~\cite{zhang2024earthgpt}} & \multirow{3}{*}{$\times$} & \multirow{3}{*}{$\times$} & \multirow{3}{*}{1005k} & Image Caption, Region Caption, VQA, Scene Classification, Visual Grounding, Object Detection, Multi-Task Conversation   \\
\multirow{3}{*}{SkyEye-968k~\cite{zhan2024skyeyegpt}} & \multirow{3}{*}{$\times$} & \multirow{3}{*}{$\times$} & \multirow{3}{*}{968k} & Image Caption, VQA, Scene Classification, Visual Grounding, UAV Video Caption, Phrase Grounding, Multi-Task Conversation   \\
\multirow{2}{*}{LHRS-Instruct~\cite{muhtar2024lhrs}} & \multirow{2}{*}{$\times$} & \multirow{2}{*}{$\times$} & \multirow{2}{*}{81.8k} & Image Caption, VQA, Scene Classification, Visual Grounding, Multi-Task Conversation  \\ \hline
\multirow{5}{*}{FIT-RS (Ours)} & \multirow{5}{*}{\checkmark} & \multirow{5}{*}{\checkmark} & \multirow{5}{*}{\textbf{1800.8k}} & \begin{minipage}[t]{\linewidth}Image Caption, Region Caption, VQA, Multi-Label Scene Classification, Object Detection, Multi-Task Conversation \newline Relation Detection, Relation Reasoning, Object Reasoning, Region-Level SGG, Image-Level SGG\end{minipage} \\
\bottomrule

\end{tabularx}
\vspace{-12pt}
\end{table}

Intelligent interpretation of remote sensing images (RSIs) is crucial and meaningful. It provides valuable information for applications ~\cite{bharatkar2013evaluation,everingham2015pascal}. Some studies collect remote sensing-specific image-text datasets~\cite{zhang2023rs5m,wang2024skyscript,mall2023remote,yuan2024chatearthnet} from open-source geographic data, incorporating basic remote sensing knowledge into models. Furthermore, existing works~\cite{hu2023rsgpt,kuckreja2023geochat,zhang2024earthgpt,zhan2024skyeyegpt,muhtar2024lhrs,pang2024h2rsvlm,RSGPT4V} propose instruction tuning datasets that encompass various downstream interpretation tasks, to obtain comprehensive Remote Sensing Large Multi-Modal Models (RSLMMs). To enhance the model's interpretation accuracy, several region-level comprehension tasks~\cite{zhang2024earthgpt,zhan2024skyeyegpt} are introduced, and some datasets design simple templates to get basic spatial relationships among targets~\cite{kuckreja2023geochat}. However, when facing complex scenes with numerous targets, they are far from performing \textbf{fine-grained} comprehension. ``Fine-grained'' refers to accurately describing the fine-grained semantic relationships among targets like ``driving in the same lane with'', rather than the simple rule-based relationships like ``on the left of''.

Considering the above issues, we construct a large-scale fine-grained instruction tuning dataset, named FIT-RS (Remote Sensing Fine-Grained Instruction Tuning), which contains \textbf{1,800,851} high-quality instruction samples, aiming at enhancing the fine-grained comprehension ability of RSLMMs. As illustrated in Table ~\ref{tab:sample_compare}, FIT-RS outperforms existing datasets in multiple aspects. Specifically, FIT-RS not only covers high-quality basic comprehension tasks, such as detailed image caption with an average length of \textbf{619.21}, but also contains several novel complex comprehension tasks. Based on the tasks in FIT-RS, we further establish the FIT-RSFG (Remote Sensing Fine-Grained) benchmark.

Additionally, we establish the first benchmark to evaluate the remote sensing relation comprehension ability of LMMs, named FIT-RSRC (Remote Sensing Relation Comprehension). Following the mode of popular benchmarks in the general field, FIT-RSRC contains multiple-choice questions and adopts the CircularEval strategy~\cite{liu2023mmbench,wang2024all}. Moreover, FIT-RSRC features high-quality distractor options based on commonsense word lists created by experts and GPT-4, which increase the diversity and fairness of the benchmark. Our contributions can be concluded as follows:

\begin{itemize}[left=0pt]
  \item  A large-scale instruction tuning dataset, FIT-RS, is proposed, which contains over 1800k samples, covering basic interpretation tasks and novel complex comprehension tasks. It aims to enhance RSLMMs' ability to understand complex scenes, particularly by improving their perception of object relations through tasks such as relation reasoning and scene graph generation.
  \item We introduce SkySenseGPT, a comprehensive RSLMM that performs excellently on public remote sensing datasets and is capable of handling various complex comprehension tasks.
  \item  We propose the first benchmark for evaluating LMMs' relation comprehension ability on remote sensing scenes, named FIT-RSRC. It includes high-quality distractor options derived from commonsense word lists, as well as unanswerable questions. Our proposed SkySenseGPT achieves an overall accuracy of 55.5\% in FIT-RSRC, surpassing existing LMMs and RSLMMs.

\end{itemize}

\section{Related Work}

\subsection{Complex Comprehension Tasks of General Large Multi-Modal Models}
Existing LMMs acquire powerful comprehension abilities from large-scale, high-quality multimodal data from natural domains. In particular, the introduction of scene graph data~\cite{krishna2017visual,xu2017scene,yang2022psg} allows general LMMs to carry out fine-grained tasks. For example, Pink~\cite{xuan2023pink} introduces the referential comprehension task, enabling models to accomplish tasks like visual relation reasoning and visual spatial reasoning. All-Seeing project v2~\cite{wang2024all} proposes a whole-image scene graph generation task and introduces CRPE benchmark to evaluate LMMs' object recognition and relation comprehension abilities. These complex comprehension tasks involve modeling and evaluating explicitly annotated relationships between targets, enhancing fine-grained understanding of the images.

\begin{figure*}[!t]
        \centering
            \includegraphics[width=\textwidth]{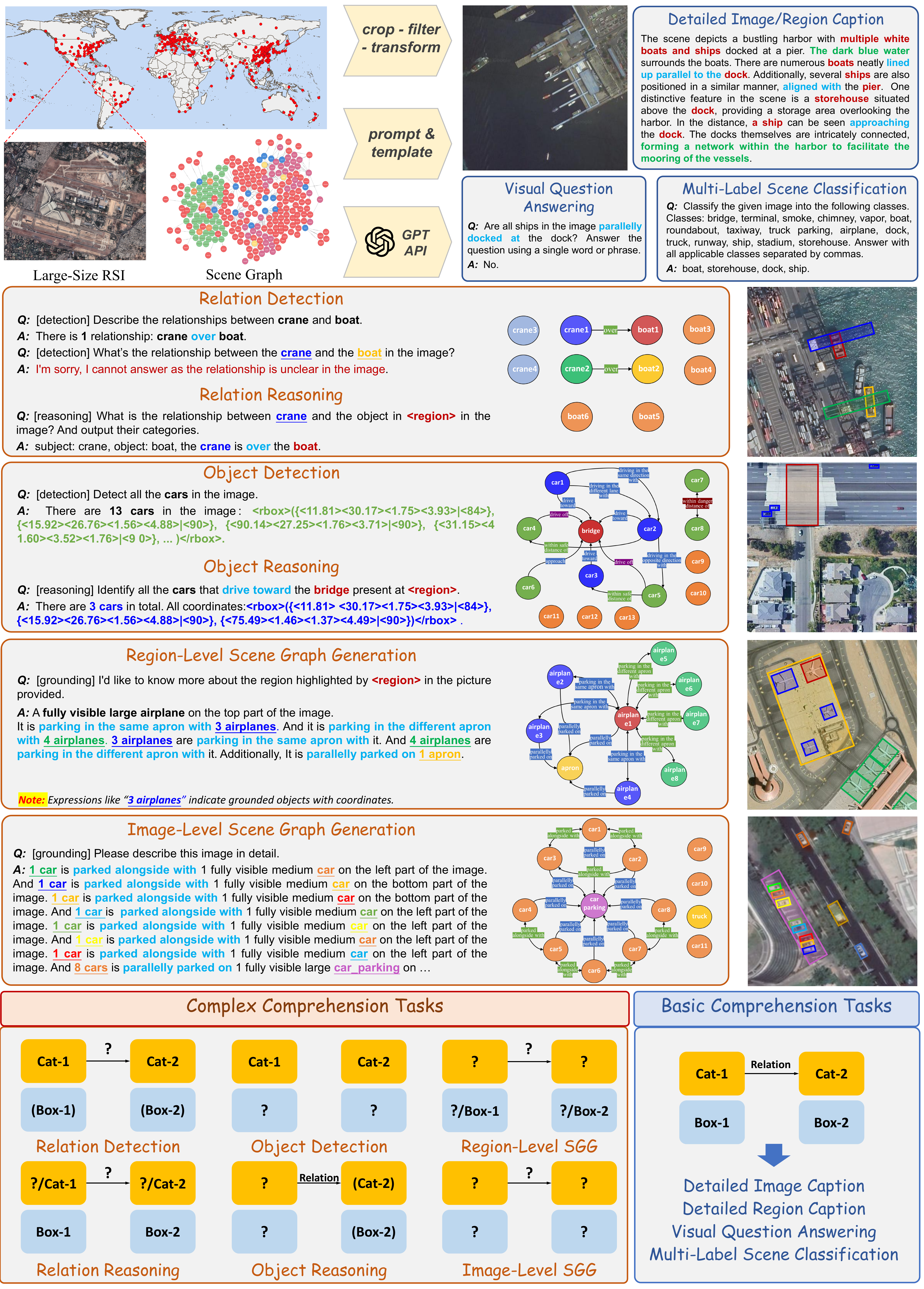}\\
           \vspace{-3pt}
	\caption{Overview of the FIT-RS dataset. The upper-right and middle parts of the figure display examples of each task. The \textbf{\uline{bold underline}} indicates the category and bounding boxes of the targets, i.e., target grounding. The below part showcases the design approach for each task, basic comprehension tasks are based on original scene graph labels, while complex comprehension tasks are formed as a set of progressively complex tasks. The ``?'' represents the content that needs to be output by the model, and the ``()'' indicates the content that will be randomly provided in the input.}
   \label{fig:tasks_definition}
\end{figure*}

\subsection{Large Multi-Modal Models in Remote Sensing}

Existing RSLMMs~\cite{hu2023rsgpt,kuckreja2023geochat,zhang2024earthgpt,zhan2024skyeyegpt,muhtar2024lhrs,pang2024h2rsvlm} have contributed to conversational understanding of RSIs, covering many basic visual-language downstream tasks. To strengthen RSLMMs' precise interpretation ability, H$^2$RSVLM~\cite{pang2024h2rsvlm} focuses on the inevitable ``hallucination'' problem and proposes the RSSA dataset to enhance the self-awareness capability of RSLMMs. Additionally, some works propose region-level comprehension tasks. For example, Geochat proposed grounded image captioning~\cite{kuckreja2023geochat}, SkyEyeGPT~\cite{zhan2024skyeyegpt} introduces the phrase grounding task. However, existing works apply simple templates to create relationships, resulting in inaccurate relationships, and can not capture complex relationships related to the targets' working or motion states. Such complex relationships are crucial for the fine-grained understanding of the scene.

\section{The FIT-RS Dataset}

Considering the rich semantic relationships among entities in complex remote sensing scenes, we aim to enhance the fine-grained comprehension capabilities of RSLMMs through instruction tuning, thus constructing the FIT-RS dataset. The overview of the FIT-RS is shown in Figure~\ref{fig:tasks_definition}. In the following sections, we will introduce the data collection and processing, the task definitions and creation in FIT-RS, and the development of the FIT-RSRC benchmark.

\subsection{Data Collection and Processing}

First, we collect large-size very-high-resolution (VHR) RSIs worldwide and manually annotate them with detailed scene graph labels, forming the STAR dataset~\cite{RSG}. These images cover multiple complex semantic scenarios (e.g., airports, dams, and ports), with the sizes ranging from 512 $\times$ 768 to 27,860 $\times$ 31,096 pixels. Our FIT-RS dataset is an extension of the STAR and contains more than 400,000 triplets and over 210,000 objects with rotated bounding boxes (RBoxes) in 1,273 RSIs. It includes 48 categories of important targets and 58 categories of high-value semantic relationships. Specifically, the semantic relationships are defined into 8 major categories: \textit{``distance warning''}, \textit{``spatial topology''}, \textit{``functional description''}, \textit{``circuit layout''}, \textit{``movement status''}, \textit{``emission status''}, \textit{``construction status''} and \textit{``parking status''}, which contain 58 subcategories. These deeply characterize the target in context. Within the scenes, diverse objectives share rich inter-class and intra-class relationships, furnishing foundational support for fine-grained scene comprehension. Due to the limited resolution of existing image-text visual encoders~\cite{radford2021learning}, we employ the common-used cropping strategy~\cite{ding2021object} to process large-size images. Using a sliding window of 512 $\times$ 512 pixels with a 100-pixel overlap, original large-size RSIs are cropped into patches. Then we filter targets and relationships using a completeness threshold, obtaining 82,532 valid image patches.



As shown in the below part of Figure \ref{fig:tasks_definition}, in the scene graph, when considering object location, the triplets can be expressed as: \textit{<subject-box, relation, object-box>}. These elements form the foundation for understanding the interactive relationships among objects. However, it is difficult for RSLMMs to directly understand the entire scene graph. To better explore the model's fine-grained understanding capabilities, we tailor a series of instruction following tasks grounded on the above fundamental elements. For \textbf{basic comprehension tasks}, we construct the instruction from all scene graph information. For \textbf{complex comprehension tasks}, we adjust the known elements in the input questions and ask the model to provide the unknown elements. This approach allows the creation of a set of progressively complex tasks, thereby fostering LMMs' fine-grained understanding of intricate remote sensing scenarios. Table \ref{tab:sample_num} shows the details of samples within the FIT-RS dataset.

\begin{table}[!h]
\caption{Details of instruction tuning samples in FIT-RS. The abbreviations and their full meanings are as follows: IC: image caption, RC: region caption, VQA: visual question answering, SC: scene classification, CC: complex comprehension, MT: multi-turn conversation.}
\label{tab:sample_num}
\renewcommand{\arraystretch}{1.3}
\centering
\setlength{\tabcolsep}{15pt}
\begin{tabular}{ccc}
\toprule
    Task Type  & Task & Sample Number \\\hline
    IC  & Detailed Image Caption & 82532 \\
    RC & Detailed Region Caption   & 90744  \\
    VQA & Visual Question Answering  & 498565  \\
    SC & Multi-Label Scene Classification   & 165074  \\
    \hline
    \multirow{6}{*}{CC}
               & Relation Detection         & 181799   \\
               & Relation Reasoning         & 124496   \\
               & Object Detection  & 140626   \\
               & Object Reasoning           & 295395   \\
               & Region-Level Scene Graph Generation   & 93199   \\
               & Image-Level Scene Graph Generation & 62945 \\ \hline
    MT & Multi-Turn/Task Conversation & 65476   \\
\bottomrule
\end{tabular}
\end{table}

\subsection{Data Construction for Basic Comprehension Tasks}
Similar to common remote sensing visual-language tasks, we construct several basic comprehension tasks. However, our distinction lies in improving the quality and richness of the instructions by making full use of fine-grained relationships.
\vspace{-3pt}
\begin{itemize}[left=0pt]
    \item \textbf{Detailed Image \& Region Caption.} Benefiting from the scene graph's high-quality detailed annotation relations among objects, we are capable of obtaining in-depth portrayals of images. The \textbf{Region Caption} task follows the subsequent region-level SGG but removes the bounding boxes. For the detailed \textbf{Image Caption} task, the generation pipeline is as follows.

    (i) First, we extract valid triplets from the patches after cropping as the foreground information. (ii) Then we use the efficient TinyLLaVA-3.1B~\cite{zhou2024tinyllava} to swiftly generate concise background descriptions for numerous RSIs. (iii) Finally, we combine them as the prompt to obtain detailed and fluent sentences using GPT-4/GPT-3.5~\cite{achiam2023gpt}. In this way, we obtain 82,532 image-caption pairs in total, with 19,874 by GPT-4 and 62,658 by GPT-3.5, the average character length of the caption generated by GPT-4 is \textbf{892.66} and the overall average length is \textbf{619.21}, significantly surpassing existing remote sensing image-text datasets (e.g., RS5M~\cite{zhang2023rs5m}: 87, SkyScripts~\cite{wang2024skyscript}: 56, HqDC-1.4M~\cite{pang2024h2rsvlm}: 369).

\item \textbf{Visual Question Answering.} Benefiting from the rich relationships among objects, we can perform more fine-grained Visual Question Answering. Compared to conventional VQA tasks, we have also introduced relation-based question answering.

Specifically, we design 4 types of questions as follows. (i) \textbf{Presence Questions.} These inquiries address the existence of a specific category. We balance positive and negative responses to avoid an overabundance of negative answers. (ii) \textbf{Comparison Questions.} We follow the existing datasets~\cite{lobry2020rsvqa} to compare the quantities of two categories. (iii) \textbf{Count Questions.} We design single-category counting questions and introduce relational constraints, asking for the number of a specific triplet in the image. (iv) \textbf{Relation Questions.} We design questions targeting the diversity of relationships. We select one \textit{subject-object} category pair to inquire about the existence of a specific relation or choose a triplet to question whether the uniqueness of that relationship is satisfied. In summary, the VQA tasks in FIT-RS are more detailed compared to existing datasets and more thoroughly test RSLMMs' fine-grained image understanding ability.

\item \textbf{Multi-Label Scene Classification.} We first manually classify original RSIs into 10 main scene categories. Subsequently, we design quantity-based filtering rules to select the primary target categories from the image patches, which serve as secondary class labels in a multi-label context. The final scene classification task contains two types of instructions: single-label scene classification and multi-label target classification.
\end{itemize}

\subsection{Data Construction for Complex Comprehension Tasks}

Fined-grained scene comprehension ability is important for RSLMMs while existing datasets only provide limited comprehension tasks, e.g., visual grounding and region caption. We further explore more challenging tasks besides the above basic tasks. Specifically, to make full use of the triplets \textit{<subject, relation, object>} in the scene graph, we set up a series of complex comprehension tasks, focusing sequentially on relation understanding, subject/object comprehension, and overall triplet interpretation, as illustrated in the lower part of Figure~\ref{fig:tasks_definition}. Next, we provide a detailed introduction to these complex comprehension tasks. 

\begin{figure*}[!h]
    \centering
        \includegraphics[width=\textwidth]{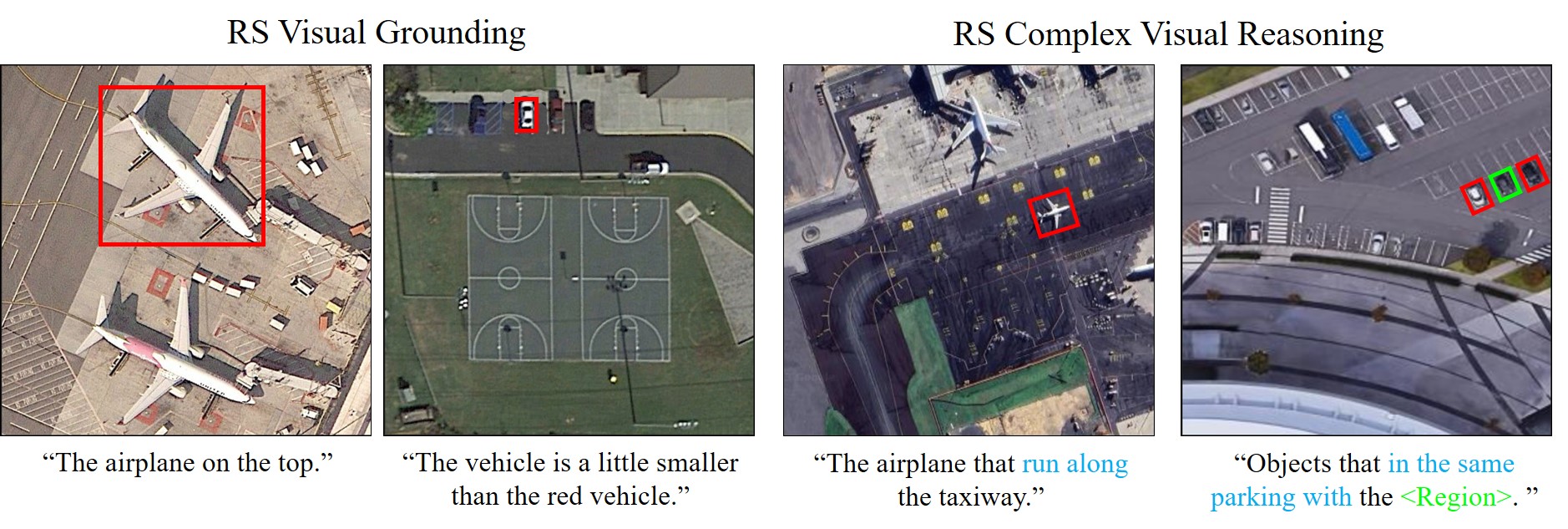}\\
       \vspace{-3pt}
\caption{Comparison between traditional remote sensing visual grounding task and our proposed fine-grained object reasoning task.}
\label{fig:task_compare}
\end{figure*}

\begin{itemize}[left=0pt]
    \item \textbf{Relation Detection \& Reasoning.} Benefiting from the precisely annotated rich semantic relations in the scene graph, we designed two tasks to enhance the RSLMM's perception of visual relations among targets. From the categories and locations of subjects and objects, we randomly select partial information as incomplete input. The Relation Detection task requires the model to output only the relations among the entities. The Relation Reasoning task additionally requires the model to output the categories of the subject and object. The relation is marked by \textit{<rel></rel>}.

    \item \textbf{Object Detection \& Reasoning.} These two tasks involve region-based understanding and outputs. Considering the precision of RBoxes~\cite{yang2021learning,zhou2022mmrotate,yang2023detecting}, we employ the ``oc'' definition to represent precise RBoxes, and use special tag \textit{<rbox>} to the location of entities in the format: \textit{<rbox>({<cx><cy><w><h>|<angle>})</rbox>}. Here, the coordinates are normalized to [0.00, 100.00] with two decimal places, and the angle is represented as an integer in degrees. (i) \textbf{Object Detection} aims for the model to detect and output all coordinates of a specific category. (ii) \textbf{Object Reasoning} is more challenging: give a specific relation and random categories and locations as the query, the model is required to output the categories and locations of all targets that meet the conditions. This task supports region input and relation queries, like ``\textit{Provide the coordinates of all airplanes that are <rel>parallelly parked on</rel> the apron present at <region> in the image.}'', assisting users in achieving more precise object interpretation. As illustrated in Figure~\ref{fig:task_compare}, our object reasoning task offers certain advantages compared to traditional remote sensing visual grounding task~\cite{zhan2023rsvg}.

    \item \textbf{Region-Level \& Image-Level Scene Graph Generation.} These two tasks are relatively the most challenging, aiming to explore the upper boundary of RSLMMs' fine-grained understanding capabilities. (i) \textbf{Region-Level SGG} requires the model to describe the target's relative size, position, and visibility based on the input area, then generate all related triplets, using ``it'' to refer to the target; (ii) \textbf{Image-Level SGG} involves directly generating scene graphs for the whole image, with isolated targets that have no relationships being appended to the answer in the format of object detection. Grounded targets are represented as \textit{<ref>target></ref><rbox>box</rbox>}. 
    
    For all the aforementioned tasks, we construct corresponding negative samples, which are instructions to refuse to answer when it comes to background images, to enhance the stability of the model. Templates for all complex comprehension tasks and negative sample responses can be found in the Appendix. Additionally, we randomly mix different tasks to create multi-turn conversation data.
    
\end{itemize}

\subsection{Evaluation of Remote Sensing Relation Comprehension}

\begin{figure*}[!th]
        \centering
            \includegraphics[width=\textwidth]{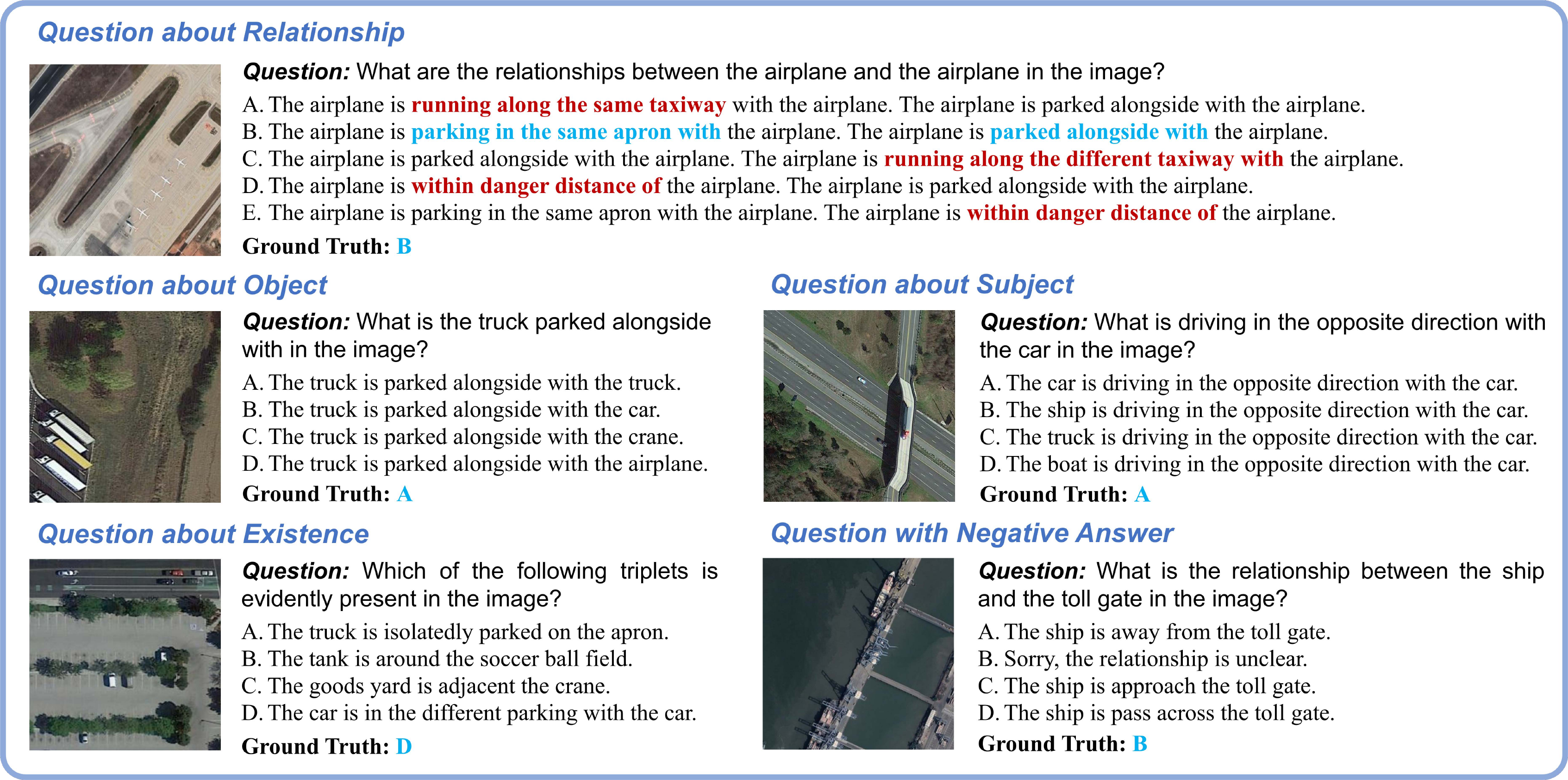}\\
	\caption{Some examples in our proposed FIT-RSRC. FIT-RSRC contains 4 types of questions and adopts the CircularEval strategy, with high-quality distractor options (from expert commonsense word lists set) and unanswerable questions. }
	\label{fig:rsrc_bench}
\end{figure*}

Given the current lack of a publicly available benchmark for comprehensive and quantitative evaluation of existing LMMs in remote sensing relation understanding, we propose the FIT-RSRC (Remote Sensing Relation Comprehension) benchmark. It is designed in the form of single-choice questions, containing four different types of questions and high-quality distractor options as in Figure \ref{fig:rsrc_bench}. Following the mainstream general benchmark, FIT-RSRC employs CircularEval~\cite{liu2023mmbench} as the evaluation strategy to enhance fairness. 

To ensure the high quality of distractors, all incorrect options are generated within the scope of a preset commonsense word list, which is created by GPT-4 in conjunction with expert manual screening. This ensures that incorrect options that can be judged solely by commonsense expressions, such as \textit{``taxiway is running along the storehouse''}, do not occur. As shown in the first example in Figure \ref{fig:rsrc_bench}, each sample in our benchmark consists of one correct option and three distractors. If the correct answer contains multiple relationships, the length of the incorrect answer will be kept equal to it. At the same time, the wrong options may incorporate correct content to improve the quality of the distractors. Moreover, to assess both the veracity of the models and their robustness against negative images, we have incorporated unanswerable answers, thus incorporating the LMMs' capability to hallucinate negative instances into our quantitative evaluation.

\section{The SkySenseGPT Model}

The architecture SkySenseGPT follows mainstream LMMs, composed of a visual encoder, i.e. CLIP-ViT-L14~\cite{radford2021learning} with a resolution of 336 $\times$ 336, a multilayer perceptron (MLP) as the multi-modal projector, and an LLM (Vicuna-v1.5~\cite{vicuna2023}). Input images are resized to 504 $\times$ 504 pixels following the Geochat~\cite{kuckreja2023geochat}. And we follow LLaVA-v1.5~\cite{liu2023improved} to get the pre-trained MLP projector. For the instruction tuning stage, we freeze the vision encoder, finetune the projector, and adopt LoRA~\cite{hu2021lora} to finetune the LLM.

We split the FIT-RS dataset into train, val, and test sets with a ratio of 6:2:2. We use the train-val set for instruction tuning, and considering the efficiency of LMM inference, we randomly sample a portion from the test set for evaluation. In this way, we obtained 1415k instruction samples for training. Furthermore, we additionally collect existing public datasets and transform them into the final instruction tuning data for training. Specifically, we collected 3 scene classification datasets (NWPU~\cite{cheng2017remote}, UCM~\cite{qu2016deep}, and RSITMD~\cite{yuan2022exploring}), 3 VQA datasets (EarthVQA~\cite{wang2024earthvqa}, Floodnet-VQA\cite{rahnemoonfar2021floodnet}, and RSVQA-LR~\cite{lobry2020rsvqa}), and 3 object detection datasets (DOTA-v2.0~\cite{ding2021object}, DIOR~\cite{cheng2022anchor}, and FAIR1M~\cite{sun2022fair1m}) to enrich the instruction dataset. This part of the additional data contains 365k samples. Detailed information can be found in the Appendix.

\vspace{-5pt}
\section{Experiments}
\vspace{-5pt}
In this section, we present the experimental results of the FIT-RSFG benchmark based on the FIT-RS dataset and the FIT-RSRC benchmark. Moreover, we conduct generalization experiments with SkySenseGPT using existing public datasets. All experiments are conducted in 4 NVIDIA A100 GPUs (40G). The specific details for all test sets can be found in the Appendix. 

\vspace{-5pt}
\subsection{Benchmarks}

We employ existing mainstream models to evaluate all the tasks in the FIT-RS dataset, creating the FIT-RSFG benchmark. To construct the FIT-RSRC benchmark, we extensively evaluate various LMMs and RSLMMs, demonstrating FIT-RSRC's reliability and ease of generalization.

\begin{table}[!h]
\caption{Result of the baseline models on the test set of each task in FIT-RS. Both zero-shot results and fine-tuning results are reported. ``ZS'' indicates the zero-shot setting and ``FT'' indicates the fine-tuning setting. ``Failed'' indicates that due to the gap between instruction formats or the model's lack of capability, the results could not be properly evaluated.}
\renewcommand{\arraystretch}{1.3}
\small
\label{tab:FIT_res}
\centering
\begin{tabular}{cc|c|c c c}
\toprule
\textbf{Task Type}     & \textbf{Task}     & \textbf{Metric}         & \thead{LLaVA1.5-7B\\(ZS)} & \thead{GeoChat\\(ZS)} & \thead{SkySenseGPT\\(FT)} \\ \hline
IC  & Detailed Image Caption                               & BLEU-1         & 15.38      & 8.79        &   27.31       \\
RC & Detailed Region Caption                              & BLEU-1         & 14.70      & 14.55       &   75.82       \\

     VQA &             VQA        & Ave Acc        & 58.59      & 53.47       &  79.76        \\
\multirow{2}{*}{SC} & \multirow{2}{*}{\thead{Multi-Label\\Scene Classification}} & Scene Acc & 56.18      & 42.27       &    82.23      \\

               &                             & Obj F1-Score   & 12.84      & 14.95       &  71.43       \\\hline

\multirow{6}{*}{CC} & Relation Detection                     & F1-Score       & Failed     & Failed      &   88.68       \\

   &  Relation   Reasoning                          & F1-Score       & Failed     & Failed      &   74.33       \\ 
&Object   Detection                          & mAP            & Failed     & 6.13        &    27.40      \\
&Object   Reasoning                         & mAP            & Failed     & Failed      &    5.71      \\
&Region-Level   SGG           & Recall         & Failed     & Failed      &  17.01        \\
&Image-Level   SGG             & Recall         & Failed     & Failed      &   9.60       \\\hline

MT & Multi-Turn/Task Conv  & GPT-eval & 2.74     & 3.64      &   6.60       \\
\bottomrule
\end{tabular}
\vspace{-5pt}
\end{table}

\subsubsection{FIT-RSFG Benchmark}

\vspace{-3pt}

To construct the FIT-RSFG benchmark, we employ LLaVA-v1.5~\cite{liu2023improved} and Geochat~\cite{kuckreja2023geochat} as our baseline models and evaluate SkySenseGPT for fine-tuning results. We adopt the common metric mAP~\cite{everingham2015pascal} with an IoU threshold of 0.25 for tasks related to target localization. For SGG tasks, we adopt the recall~\cite{Motif,Vctree} and mean recall metrics~\cite{T1,T2}. We employ GPT-3.5 to get GPT-eval scores, which range from 0 to 10. The results are shown in Table~\ref{tab:FIT_res}. After fine-tuning, the model can handle complex tasks, showing a clear improvement compared to the baseline methods. The detailed results can be found in the Appendix.

\subsubsection{FIT-RSRC Benchmark}
\vspace{-3pt}
Due to the conciseness of the single-choice format, we evaluate several general-domain LMMs, including LLaVA-1.5-7B~\cite{liu2023improved}, LLaVA-HR-7B~\cite{luo2024feast}, and TinyLLaVA ~\cite{zhou2024tinyllava} that we used in constructing image descriptions. We also evaluate currently open-sourced RSLMM like GeoChat~\cite{kuckreja2023geochat} on the FIT-RSRC.

\begin{table}[!h]
\small
\caption{Accuracy of LMMs on the FIT-RSRC, employing CircularEval strategy. ``TL'' stands for cross-task transfer learning, which indicates that the model has been exposed to images from the same source training set, but has not encountered this specific type of task before.}
\renewcommand{\arraystretch}{1.2}
\centering
\label{tab:RSRC_res}
\begin{tabular}{c|c|c|c|c|c|c}
\toprule
\textbf{Method} & \textbf{Setting} & \textbf{Subject} & \textbf{Object} & \textbf{Relation} & \textbf{Existence} & \textbf{Ave. Acc} \\ \hline
LLaVA-1.5-7B~\cite{liu2023improved} &  \multirow{4}{*}{ZS} & 6.4              & 11.2            & 21.0                & 7.8            & 11.6             \\
TinyLLaVA ~\cite{zhou2024tinyllava} &   & 22.0              & 18.0            & 12.0                & 55.6            & 26.9             \\
LLaVA-HR-7B~\cite{luo2024feast}     &   & 7.8              & 13.4            & 32.6              & 47.8           & 25.4             \\
GeoChat~\cite{kuckreja2023geochat}    &     & 7.6              & 8.6             & 25.6              & 45.6           & 21.9            \\ \hline
SkySenseGPT  & TL   & 25.2             & 54.0            & 50.4            & 92.2           & 55.5  \\
\bottomrule
\end{tabular}

\end{table}

The results are shown in Table~\ref{tab:RSRC_res}. Based on the results, existing LMMs generally showcase a good ability to identify the presence or absence of triplets, indicating their basic capability to recognize objects in remote sensing scenes. However, when it comes to fine-grained understanding related to object relationships, the performance of existing models is generally poor. It is worth noting that LLaVA-HR achieves high accuracy in the zero-shot setting, indicating that high-resolution input is significant for the fine-grained understanding of RSIs. Despite being set up for cross-task transfer learning, SkySenseGPT demonstrates a better ability to recognize relationships than the other models.


\subsection{Generalization Experiments}

\begin{table}[!h]
  \small
  \centering
  \setlength{\tabcolsep}{4.5pt}
  \renewcommand{\arraystretch}{1.4}
  \caption{Comparion of RSLMMs on public datasets. For the VQA datasets, similar to prior works~\cite{hu2023rsgpt,kuckreja2023geochat,pang2024h2rsvlm}, we omit area and count questions during evaluation.}
  \label{tab:public_res}
    \begin{tabular}{c|c|c|c|cccc}
    \toprule
    \arrayrulecolor{black}
    \textbf{Task}   & \textbf{Setting} & \textbf{Dataset}    & \textbf{Metric} & \thead{Geo-\\Chat~\cite{kuckreja2023geochat}} & \thead{LHRS-\\Bot~\cite{muhtar2024lhrs}} & \thead{H$^2$RS\\VLM~\cite{pang2024h2rsvlm}}  & \thead{SkySense\\GPT} \\ \hline
    \multirow{4}{*}{\thead{Scene\\Classification}} 
    & \multirow{4}{*}{ZS}     & SIRI-WHU      & Acc  &  66.63 &  62.66 &  \uline{68.50}   &  \textbf{74.75}  \\
    &     & AID    & Acc  &  72.03  &  \uline{91.26} &  89.33   & \textbf{92.25}   \\
         &    & WHU-RS19 & Acc & 86.47   &  93.17 &  \uline{97.00}  &  \textbf{97.02}   \\                              
       &  & AID-multi & F1-Score &   46.55 & -  &  -    &  \textbf{47.97}  \\  \hline
    \multirow{7}{*}{VQA}     & \multirow{3}{*}{ZS}   &  \multirow{3}{*}{RSVQA-HR}
    & Pre. Acc  & 58.45 &  -  & \uline{65.00}    & \textbf{69.14}     \\
    & & & Comp. Acc  & 83.19  &  -  & \uline{83.70}    & \textbf{84.14}    \\
    & & & Avg. Acc  & 70.82 &  -  & \uline{74.35}    & \textbf{76.64}     \\\cline{2-8}
    &\multirow{4}{*}{FT}  & \multirow{4}{*}{RSVQA-LR}   & Rural. Acc & \uline{91.09}       &  89.07    & 88.00   &      \textbf{95.00}     \\ 
   && & Pre. Acc & \uline{90.33}   & 88.51     & 89.58       &       \textbf{91.07}     \\ 
   && & Comp. Acc & \textbf{94.00}  & 90.00     & 89.79         &       \uline{92.00}     \\ 
   && & Avg. Acc  & \uline{91.81} & 89.19     & 89.12        &       \textbf{92.69}     \\
    \bottomrule
    \end{tabular}%
    
\end{table}

We compare our SkySenseGPT with state-of-the-art RSLMMs in public datasets as in Table~\ref{tab:public_res}. For scene classification, we choose four representative datasets SIRI-WHU~\cite{zhu2016bag}, AID~\cite{xia2017aid}, WHU-RS19~\cite{dai2010satellite} and AID-Multilabel~\cite{hua2019relation} for zero-shot generalization experiments. For VQA, we choose two representative datasets: RSVQA-HR and RSVQA-LR~\cite{lobry2020rsvqa}. In the scene classification task, SkySenseGPT surpasses the current RSLMMs, especially exceeding state-of-the-art H$^2$RSVLM~\cite{pang2024h2rsvlm} by \textbf{6.25 \% }on the SIRI-WHU dataset. And in the VQA task, our model achieves high accuracy under both low and high resolutions, indicating that it can serve complex fine-grained tasks and possess strong basic comprehension capabilities for foundational tasks.


\section{Conclusion}
\vspace{-4pt}
In this paper, we construct FIT-RS, a large-scale instruction tuning dataset aimed at enhancing the fine-grained comprehension capability of RSLMMs. It covers various tasks and features high-quality corpora, including basic and novel complex tasks. Based on the FIT-RS, we construct the FIT-RSFG benchmark. Furthermore, we construct the first benchmark, FIT-RSRC, for evaluating LMMs' ability in remote sensing relationship comprehension. By combining FIT-RS with other multi-modal corpora, we further introduce SkySenseGPT, which excels in different granular tasks and surpasses current RSLMMs in public datasets. As there is still room for improvement in challenging complex tasks, we hope that FIT-RS can contribute to building more powerful RSLMMs.

\bibliographystyle{unsrt}
\bibliography{main}

\newpage
\appendix

\section*{Appendix}
\setcounter{figure}{0}
\setcounter{table}{0}
\renewcommand{\thetable}{A\arabic{table}}
\renewcommand{\thefigure}{A\arabic{figure}}

\section{Overview}

The FIT-RS dataset, data generation scripts, and model weight of SkySenseGPT will be open-sourced at \url{https://github.com/Luo-Z13/SkySenseGPT}, the DOI of the data repository is \href{https://huggingface.co/datasets/ll-13/FIT-RS}{10.57967/hf/2529}. In this Appendix, we provide a detailed description of the templates and prompts used to create various instruction tuning tasks in the FIT-RS dataset. We also provide specific details about our constructed datasets and the public datasets used for training and testing. Additionally, we present detailed experimental results and comprehensive visualization results.

\section{The FIT-RS Dataset}

We provide a detailed explanation of the specific question templates and prompts used for different tasks in the FIT-RS dataset, and their specific formats are shown in Table~\ref{tab:task_format}. During data preprocessing, as we employ the conventional cropping strategy to cut the original large-size remote sensing images (RSIs) into small patches, we set the completeness threshold, i.e., the intersection over foreground (iof) threshold, as 0.3 for detailed image caption task to provide comprehensive descriptions, and 0.55 for other tasks to ensure the reliability of target-related corpus.

\begin{table}[!h]
\renewcommand{\arraystretch}{1.3}
\centering
\caption{Formats of instruction following tasks in the FIT-RS. The abbreviations and their full meanings are as follows: IC: image caption, RC: region caption, VQA: visual question answering, SC: scene classification, CC: complex comprehension, MT: multi-turn conversation.}
\begin{tabular}{ccc}
\toprule
    Task Type  & Task & Specific Format \\\hline
    IC  & Detailed Image Caption & - \\
    RC & Detailed Region Caption   & -   \\
    VQA & Visual Question Answering  & Answer in one word or a short phrase.  \\
    \multirow{1}{*}{SC} & \thead{Multi-Label \\Scene Classification}   & \makecell{Classify the given image in one of the following \\ classes. / Answer in one word or a short phrase. \\ Classify the given image into the following \\  classes. / Answer with all applicable classes \\ separated by commas.} \\ 
    \hline
    \multirow{6}{*}{CC}
               & Relation Detection    & [detection]   \\
               & Relation Reasoning    & [reasoning]   \\
               & Object Detection    & [detection]   \\
               & Object Reasoning    & [reasoning]   \\
               & Region-Level SGG   & [grounding]  \\
               & Image-Level SGG & [grounding] \\ \hline
    MT & Multi-Turn/Task Conversation & -   \\
\bottomrule
\end{tabular}
\vspace{4pt}
\label{tab:task_format}
\end{table}

\subsection{Detailed Image Caption}

\textbf{Foreground annotation generation.} For each image, the annotations of scene graph generation (SGG), i.e., the object annotations and triplets, are transformed into sentences as a part of the prompt of GPT-4/GPT-3.5~\cite{achiam2023gpt} through automated scripts. The object detection annotations are arranged in a format similar to LLaVA~\cite{liu2024visual}, with the difference being that we employ rotated bounding boxes. For the triplet annotations, we do not retain objects' ID numbers, as they are unordered and would inevitably interfere with the final descriptions generated by the GPT. Additionally, to control the final prompt length to avoid exceeding the GPT-4/GPT-3.5 token limit, identical triplets are combined into one. For example, \textit{`crane over ship, crane over ship, crane over ship'} are transformed into \textit{`3 cranes over ship'}, hence avoiding subsequent generation of incomplete scene descriptions.

\textbf{Background description generation.} Owing to the absence of annotations for natural landforms and buildings in the original scene graph annotations, we employ existing LMMs to generate concise yet precise background descriptions. Specifically, using the powerful and efficient TinyLLaVA-3.1B~\cite{zhou2024tinyllava}, we can swiftly generate concise background descriptions for a large volume of RSIs at approximately 1.1 seconds per image. Despite TinyLLaVA facing hallucination issues when directly describing man-made artifacts within RSIs, it can accurately describe the main background elements like water bodies and vegetation. The prompt used for TinyLLaVA is shown in the upper part of Table~\ref{tab:image_caption_prompt}.  After producing background descriptions for 85,000 RSIs each with a resolution of 512$\times$512 pixels, we manually conducted a random review of 2,000 images, solidifying the background caption's fundamental accuracy.

\vspace{-3pt}

\textbf{Detailed caption generation.} We define a system prompt and combine the background and foreground descriptions as the user prompt. Then they are combined to fed into the GPT-4/GPT-3.5 API to obtain fluent, detailed, and comprehensive image descriptions. The prompt and information format we used is illustrated in Table~\ref{tab:image_caption_prompt}. The prompt for generating the background description and final detailed image caption is in Table~\ref{tab:image_caption_instruction}. 

Furthermore, we present examples in Table~\ref{tab:image_caption_example} and Table~\ref{tab:image_caption_example2}, demonstrating how the combination of rich, detailed foreground information and concise, precise background information enables us to generate a thorough description of the entire image using the GPT API. We can observe that the intricate relationships among objects play a crucial role in interpreting object states and gaining a deep understanding of remote sensing scenes.

\begin{table*}[h!]\centering
 \caption{For each query, we illustrate the prompt construction process for GPT3.5/GPT-4 to collect \VarSty{ query[`response']}  from \VarSty{ query[`context']}, using few-shot in-context-learning, where examples are from \VarSty{fewshot\_samples}, each example including input \VarSty{sample[`context']} and output \VarSty{sample[`response']}. Note that \VarSty{messages} is the final prompt. The \VarSty{ query[`context']} is composed of three parts: \textbf{Background Caption}, \textbf{Objects} and \textbf{Relationships}, please see examples in Table~\ref{tab:image_caption_example} and Table~\ref{tab:image_caption_example2} for details.}
 \vspace{10pt}
\begin{minipage}{0.99\columnwidth}\vspace{0mm}    \centering
\begin{tcolorbox} 
    \centering
     \hspace{-6mm}
    \begin{tabular}{p{0.99\columnwidth}}
\begin{minipage}{0.99\columnwidth}\vspace{0mm}

\VarSty{messages} = [
            \{\var{"role":"system", "content":} f"""You are an AI visual assistant that can analyze a single image. You receive a background caption describing the same image you are observing. In addition, specific object locations and relationships within the image are given, along with detailed coordinates. These coordinates are in the form of rotated bounding boxes, represented as (x1, y1, x2, y2, x3, y3, x4, y4). These values correspond to the 4 corner points of a rbox. \\
            
            Using the provided caption, bounding box and relation information, describe the scene in a detailed manner. Instead of directly mentioning the bounding box coordinates, utilize this data to explain the scene using natural language. {\bf Include details like object counts, position of the objects, and relationships between the objects.} \\
            
            When using the information from the caption and coordinates, directly explain the scene, and do not mention that the information source is the caption or the bounding box, avoid mentioning the exact value of coordinates and `iof'. Avoid subjective, emotionally charged speculation and description. {\bf Be aware that relationships might be mutual, so do not determine the number of targets based on relationships.} Always answer as if you are directly looking at the image. Don't make your answer too long."""\}\\
        ]
    \For{ \VarSty{sample} in \VarSty{fewshot\_samples}}{
         \var{\VarSty{messages}.append(\{"role":"user", "content":\VarSty{sample[`context']}\})} \; \\
         \var{\VarSty{messages}.append(\{"role":"assistant", "content":\VarSty{sample[`response']}\} ) } \;
         }  
    \var{\VarSty{messages}.append(\{"role":"user", "content":`\textbackslash  n'.join(\VarSty{query})\})}
\end{minipage}
    \end{tabular}
\end{tcolorbox}

\vspace{-2mm}
\label{tab:image_caption_prompt}
\end{minipage}
\end{table*}

\clearpage

\begin{table*}[!h]\centering
\caption{The list of instructions for detailed image caption.
}
\centering
\begin{minipage}{\linewidth}
\begin{tcolorbox} 
\begin{tabular}{p{\linewidth}}
    \begin{itemize}[left=0pt]
    \vspace{-5pt}
        \item "Describe this image in detail."
        \item "Provide a detailed description of the given image."
        \item "Analyze the image in a comprehensive and detailed manner."
        \item "Give a thorough description of the image."
        \item "Provide an in-depth analysis of the image."
        \item "Describe the elements present in this image in detail."
        \item "Give a detailed interpretation of the image."
        \item "Analyze and describe the image in detail."
        \item "Provide a comprehensive description of what you see in the image."
        \item "Describe the image thoroughly, including all the elements you can identify."
        \item "Give an extensive description of the image."
        \item "Provide a detailed explanation of the image's content."
        \item "Describe in detail what you observe in the image."
        \item "Give a comprehensive analysis of the image."
        \item "Elaborate on the details you can observe in this image."
        \item "Provide an in-depth description of every element in the image."
        \item "Detail the contents of the image."
        \item "Analyze and describe every detail you can identify in the image."
        \item "Give a detailed account of the image's content."
    \end{itemize}

\end{tabular}
\end{tcolorbox}
\end{minipage}

\label{tab:image_caption_instruction}
\end{table*}

\subsection{Detailed Region Caption} 

For the detailed regional caption task, we first create a basic description of the target in question, which includes basic location, basic size, visibility, and location representation. Then describe all triplets related to it.

\textbf{Basic Location.} The image is initially divided into 9 basic regions. Then, the target's location is determined based on the position of its center point (the center of the rotated bounding box) within these regions. The possible locations are ``center'', ``top left'', ``top right'', ``bottom left'', ``bottom right'', ``top'', ``bottom'', ``left'', and ``right''.

\textbf{Basic Size.} According to the fundamental definitions for small target detection in RSIs~\cite{cheng2023towards}, targets are categorized based on threshold values of 16 $\times$ 16, 64 $\times$ 64, and 256 $\times$ 256 pixels, into ``small'', ``medium'', ``large'', and ``giant''.

\textbf{Basic Visibility.} Based on the targets' iof value after cropping, we classify them as ``barely visible'' when iof value < 0.4, ``partially visible" when 0.4 < iof value < 0.6, ``clearly visible" when 0.6 < iof value < 1.0, and ``fully visible" when iof value equals 1.0. It is important to note that when creating region-related tasks, the rotated bounding boxes may include some background. To reduce the introduction of noise, we set a relatively high iof threshold (0.55). Consequently, the targets involved in subsequent complex comprehension tasks are mostly partially visible.

\begin{table*}[h!]\centering
\caption{One example to illustrate the combined background and foreground descriptions, as well as the generated final detailed description.}
\vspace{5pt}
\begin{minipage}{1.0\columnwidth}\vspace{0mm}    \centering
\begin{tcolorbox} 
    \centering
      \footnotesize
    \begin{tabular}{p{0.97\columnwidth} c}

& \hspace{-3.7cm} \vspace{-0.4cm}
\multirow{5}{*}{ \includegraphics[height=2.8cm]{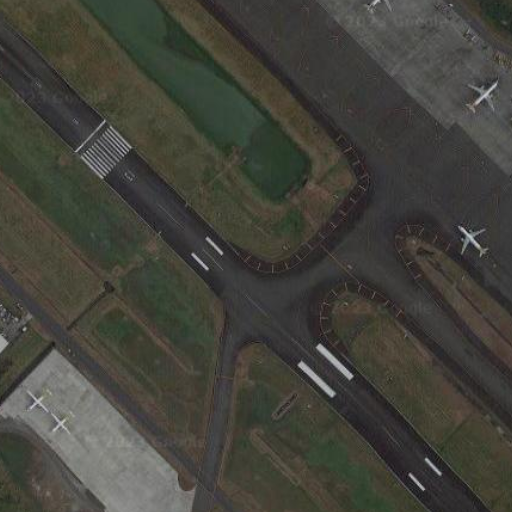} }
\\
\VarSty{ {\bf Background Caption:} } &\\
The airport runway is surrounded by grass. &\\
\vspace{0.1cm}
\VarSty{ {\bf Objects:} } & \\
apron: [10.16, 67.77, 40.62, 97.85, 29.49, 110.94, -1.76, 80.66], iof: 0.85 & \\ taxiway: [73.63, 31.84, 50.0, 56.25, 57.81, 63.67, 81.64, 38.87], iof: 1.0 & \\ airplane: [88.09, 46.29, 91.99, 42.19, 96.88, 47.66, 92.19, 51.56], iof: 1.0& \\ airplane: [93.75, 13.67, 98.63, 18.55, 93.75, 22.66, 89.26, 18.36], iof: 1.0& \\ airplane: [11.91, 86.91, 8.59, 83.2, 12.89, 78.32, 16.6, 81.84], iof: 1.0& \\ airplane: [7.23, 81.64, 3.91, 78.52, 7.03, 75.0, 10.55, 77.93], iof: 1.0 & \\ taxiway: [44.73, 58.4, 34.57, 117.77, 40.23, 117.77, 49.02, 59.38], iof: 0.67& \\ 

\VarSty{ {\bf Relationships:} } & \\
1 taxiway through apron, 2 airplanes parallelly parked on apron, 1 airplane parking in the same apron with airplane, 1 airplane parked alongside with airplane. & \\

    \hrulefill & \\
   \VarSty{ {\bf Final Image Caption:} } & \\
In the image, we observe an airport scene featuring a runway surrounded by lush grass, illustrating a \textbf{typical serene airport environment}. The scene includes various key components such as \textbf{an apron, taxiways, and airplanes}, arranged in a coordinated manner. The \textcolor{red}{\textbf{apron}}, which serves as a parking area for airplanes, is prominently positioned and appears to have ample space. It is somewhat rectangular but with an irregular shape, indicating how it accommodates the specific layout and needs of the airport. There are \textcolor{brown}{\textbf{two taxiways}} visible in the scene. One taxiway is \textcolor{blue}{\textbf{located away from}} the apron, well-defined and appears to be facilitating the movement of aircraft between the runway and the apron. This taxiway has a clear, well-maintained pathway, indicating regular usage. The other taxiway \textcolor{blue}{\textbf{is closer to the apron}} with a more complex shape, possibly due to the terrain or specific design requirements of the airport. Despite its complexity, it plays a crucial role in aircraft navigation on the ground.  \textcolor{green}{\textbf{Four airplanes}} are spotted within the scene, each at different positions but primarily associated with the apron area. Two of these airplanes are \textcolor{blue}{\textbf{parallelly parked on}} the apron, showcasing an organized and efficient use of space. These airplanes are aligned in such a way that maximizes the apron's capacity while ensuring safety and ease of movement for ground vehicles and personnel. The other two airplanes, although individually mentioned, seem to be \textcolor{blue}{\textbf{parked in close proximity to each other}} on the same apron, hinting at a busy operational day at the airport with multiple aircraft being serviced or prepared for flights. The arrangement of these airplanes alongside each other suggests a well-coordinated effort to manage space and operational activities on the apron efficiently. Overall, the layout and operational dynamics of the airport components in the image -- from the design of the taxiways facilitating movement to the organized parking of airplanes on the apron -- \textbf{display a well-managed and functional airport environment conducive to safe and efficient air travel operations}. &
    \end{tabular}
\end{tcolorbox}
\vspace{-2mm}
    \label{tab:image_caption_example}
\end{minipage}
\end{table*}
\clearpage

\begin{table*}[t!]\centering
\begin{minipage}{1.0\columnwidth}\vspace{0mm}    \centering
\caption{One example to illustrate the combined background and foreground descriptions, as well as the generated final detailed description.}
\begin{tcolorbox} 
    \centering
      \footnotesize
    \begin{tabular}{p{0.97\columnwidth} c}

& \hspace{-3.7cm} \vspace{-0.4cm}
\multirow{5}{*}{ \includegraphics[height=2.8cm]{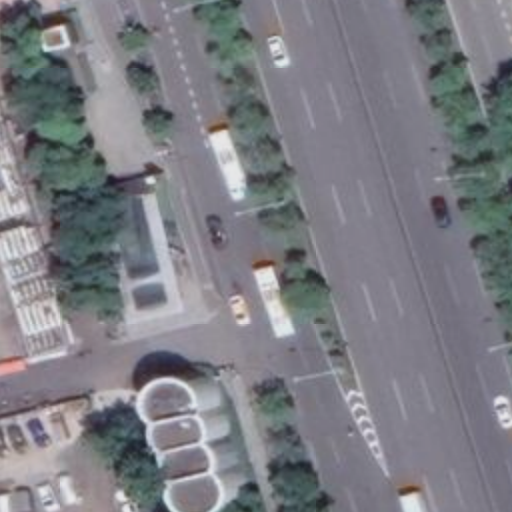} }
\\
\VarSty{ {\bf Background Caption:} } &\\
The image shows a city street with a lot of trees and buildings. &\\
\vspace{0.1cm}
\VarSty{ {\bf Objects:} } & \\
car: [57.23, 12.89, 54.88, 6.25, 51.56, 7.42, 53.91, 14.06], iof: 1.0& \\ car: [95.7, 77.73, 98.05, 84.18, 101.56, 83.01, 99.22, 76.56], iof: 0.85& \\ car: [83.59, 38.09, 85.74, 44.92, 88.67, 44.14, 86.72, 37.3], iof: 1.0& \\ car: [15.23, 98.44, 13.87, 92.38, 11.13, 92.97, 12.5, 99.02], iof: 1.0& \\ car: [12.11, 99.61, 9.57, 94.14, 6.84, 95.31, 9.38, 100.78], iof: 0.95& \\ car: [6.84, 101.76, 5.86, 95.12, 2.93, 95.51, 3.91, 102.34], iof: 0.69& \\ car: [10.55, 86.91, 7.23, 81.05, 4.49, 82.42, 7.81, 88.28], iof: 1.0& \\ car: [9.57, 80.47, 11.13, 87.11, 14.06, 86.33, 12.5, 79.69], iof: 1.0& \\ car: [6.25, 88.09, 3.91, 82.23, 1.17, 83.2, 3.32, 89.06], iof: 1.0& \\ car: [24.02, 101.17, 27.73, 100.59, 27.34, 98.05, 23.63, 98.63], iof: 0.65& \\ car: [22.85, 98.24, 25.98, 97.66, 25.59, 95.12, 22.27, 95.7], iof: 1.0& \\ car: [49.22, 63.28, 47.27, 57.03, 44.34, 57.81, 46.29, 64.26], iof: 1.0& \\ car: [44.73, 48.44, 42.58, 41.6, 39.84, 42.58, 41.99, 49.22], iof: 1.0& \\ truck: [58.01, 65.43, 52.73, 51.37, 48.83, 52.93, 54.3, 66.99], iof: 1.0& \\ truck: [49.02, 39.06, 44.34, 24.61, 40.43, 25.98, 45.12, 40.23], iof: 1.0

\VarSty{ {\bf Relationships:} } & \\
3 cars driving in the same direction with car, 3 cars driving in the same lane with car, 3 cars within safe distance of car, 7 cars parked alongside with car, 2 trucks driving in the same direction with truck, 2 trucks driving in the same lane with truck, 2 trucks within safe distance of truck. & \\

    \hrulefill & \\
   \VarSty{ {\bf Final Image Caption:} } & \\
The scene depicts a bustling city \textcolor{brown}{\textbf{street}} graced with an abundance of trees and buildings. Along this street, numerous \textcolor{green}{\textbf{vehicles}} are present, showcasing a typical vibrant urban environment. Specifically, there is a variety of \textcolor{red}{\textbf{cars}} in different states of motion and rest. Some cars are on the move; a pair of these are \textcolor{blue}{\textbf{driving in the same direction and sharing the same lane}}, maintaining a safe distance from one another. This indicates a flow of traffic on the street. Additionally, there are cars that appear to be \textcolor{blue}{\textbf{parked on the side of this street}}. A sequence of these vehicles is aligned, suggesting a designated parking area. These parked cars are \textcolor{blue}{\textbf{situated alongside each other}}, adhering to what seems to be an orderly arrangement according to parking protocols.  Moreover, two trucks are part of the street's vehicle ecosystem. They are larger than the other vehicles present and are also in motion, following the traffic pattern. These trucks are \textcolor{blue}{\textbf{driving in the same direction, staying within the same lane, and ensuring there is a safe distance between them}}, which points to the drivers' adherence to road safety measures.  The visibility of each vehicle in the street scene varies, with \textbf{some cars fully visible}, indicating clear, unobstructed views, while others have varying degrees of partial visibility, which could be due to their placement behind other objects in the environment or the viewing angle. \textbf{This typical urban snapshot represents the organized chaos of city traffic, where various vehicles share the road, adhere to driving regulations, and coexist within the street's vibrant landscape.} &
    \end{tabular}
\end{tcolorbox}
\vspace{-2mm}

    \label{tab:image_caption_example2}
\end{minipage}
\end{table*}
\clearpage

After obtaining the basic descriptions above, they can be combined to describe the user-specified region, i.e., the input of the region caption task. The combined description can translate the basic category, size, position, and visibility of a target, such as ``A fully visible medium crane on the bottom right part of the image.'' Next, we search for all triples related to the target in the scene graph annotation and convert these into natural language format, referring to the target as ``it'' to form a coherent and comprehensive region description. The list of question instructions can be found in Table~\ref{tab:region_caption_instruction}.

\begin{table*}[h!]
\caption{The list of instructions for detailed region caption.
}
\centering    
\begin{minipage}{\linewidth}
\begin{tcolorbox} 
\begin{tabular}{p{\linewidth}}
\begin{itemize}[left=0pt]
\vspace{-0.5em}
    \item "Can you provide me with a detailed description of the region in the image marked by <region>?"
    \item "I'm curious about the region represented by <region> in the picture. Could you describe it in detail?"
    \item "What can you tell me about the region indicated by <region> in the image?"
    \item "I'd like to know more about the area in the photo labeled <region>. Can you give me a detailed description?"
    \item "Could you describe the region shown as <region> in the picture in great detail?"
    \item "What details can you give me about the region outlined by <region> in the photo?"
    \item "Please provide me with a comprehensive description of the region marked with <region> in the image."
    \item "Can you give me a detailed account of the region labeled as <region> in the picture?"
    \item "I'm interested in learning more about the region represented by <region> in the photo. Can you describe it in detail?"
    \item "What is the region outlined by <region> in the picture like? Could you give me a detailed description?"
    \item "Can you provide me with a detailed description of the region in the picture marked by <region>, please?"
    \item "Please provide a detailed description of the region <region> in the given image."
    \item "What can you tell me about the region indicated by <region> in the image, exactly?"
    \item "I'd like to know more about the area in the photo labeled <region>, please. Can you give me a detailed description?"
    \item "Could you describe the region shown as <region> in the picture in great detail, please?"
    \item "Please provide me with a comprehensive description of the region marked with <region> in the image, please."
    \item "Can you give me a detailed account of the region labeled as <region> in the picture, please?"
    \item "I'm interested in learning more about the region represented by <region> in the photo. Can you describe it in detail, please?"
    \item "What is the region outlined by <region> in the picture like, please? Could you give me a detailed description?"
    \item "Please describe the region <region> in the image in detail."
    \item "Can you offer a thorough analysis of the region <region> in the image?"
    \item "Could you elaborate on the region highlighted by <region> in the picture provided?"
    \item "Please share more information about the zone emphasized with <region> in the photo."
\end{itemize}

\end{tabular}
\end{tcolorbox}
\end{minipage}
\label{tab:region_caption_instruction}
\end{table*}

\subsection{Visual Question Answering}
We meticulously designed comprehensive and diverse question templates for VQA, as illustrated in Table~\ref{tab:vqa_template}. For each type of question, we randomly select one question from the templates and then randomly choose corresponding categories or relationships based on the scene graph annotations of the image. These fill in the ``<>" in the template, generating the final question and creating the corresponding answer. Below, we outline the construction method for different types of questions.

\begin{table*}[!h]

\centering
\caption{To generate VQA questions, templates iterate through various types of image annotations, such as (Category 1, Category 2) or (Category 1, Relationship). For each type of question, we randomly select one template and fill it with the content currently being iterated.
}
\vspace{3pt}
\begin{minipage}{\linewidth}
\begin{tcolorbox} 
\begin{tabular}{p{0.97\columnwidth} c}
\VarSty{\bf Presence Question:}\vspace{0.1cm}\\

Type 1 "Is there a/an <cat name> in the image?" &\\
Type 2 "Does the image contain a/an <cat name>?" &\\
Type 3 "Can a/an <cat name> be found in the image?"
&\\[0.2cm]
\VarSty{\bf Compare Question:} \vspace{0.1cm} &\\

Type 1 "Are there fewer <cat1 name> than <cat2 name> in the image?"&\\
Type 2 "Is the number of <cat1 name> greater than the number of <cat2 name>?"&\\
Type 3 "Does the image contain the same number of <cat1 name> and <cat2 name>?"
&\\[0.2cm]

\VarSty{\bf Count Question:} \vspace{0.1cm} &\\
Type 1 "How many <cat name>s are there in the image?" &\\
Type 2 "What is the number of <cat name> in the image?"&\\
Type 3 "What is the amount of <cat name> in the image?" &\\
Type 4 "What is the count of <subj cat name>s in the image that are <rel name> a/an <obj cat name>?" &\\
Type 5 "What is the amount of <subj cat name> in the image are <rel name> a/an <obj cat name>?"&\\
Type 6 "How many <subj cat name>s in the image are <rel name> a/an <obj cat name>?" &\\
&\\[0.1cm]

\VarSty{\bf Relation Question:} \vspace{0.1cm} &\\
Type 1 "Are all <subj cat name>s in the image <rel name> a/an <obj cat name>?" &\\
Type 2 "Does the image show all <subj cat name>s <rel name> a/an <obj cat name>?" &\\
Type 3 "Is every <subj cat name> in the image <rel name> a/an <obj cat name>?"
&\\[0.2cm]



\end{tabular}
\end{tcolorbox}
\end{minipage}

\label{tab:vqa_template}
\end{table*}

\textbf{Presence Questions.} We generate questions to determine the presence of each object category in the image. For categories that are present, we create `Yes' questions confirming their presence. For absent categories, we generate `No' questions indicating their absence. We aim to balance the number of `Yes' and `No' questions generated, ensuring each category is adequately represented in the questioning process.

\textbf{Comparison Questions.} We generate questions to compare the quantities of different object categories in the image. For each pair of present categories, we create questions that ask whether one category has more, fewer, or an equal number of instances compared to the other. The questions are structured to assess the relative quantities of the identified categories in the image, providing insights into their comparative presence. Each question is accompanied by an answer ('Yes' or 'No') based on the actual counts of the categories in the image. The goal is to provide a balanced assessment of the relative quantities of different object categories present in the scene.

\textbf{Count Questions.} We generate questions to inquire about the number of specific entity categories and relationships present in the image. For each object category detected in the image, we formulate questions asking for the count or number of instances. Each question is paired with its corresponding numeric answer, reflecting the actual count of objects in the image. Additionally, we select from relationships between entities and create questions focusing on specific relationships. These questions inquire about the number of instances where a subject-object pair is connected by a specified relationship. Each question is answered based on the computed count of subject-object relationships that match the specified criteria. The aim is to provide a detailed comprehension of both the quantity of individual object categories and the relationships between them in the image.

\textbf{Relation Questions.} We generate questions to investigate the existence and specificity of relationships between categories of objects detected in the image. When effective relationship triplets are identified, we proceed to formulate questions focusing on the presence and consistency of these relationships. These questions inquire whether all instances of a particular object category are involved in a specified relationship with another object category in the image. Answers to these questions depend on whether all instances of the subject category are consistently connected by the specified relationship to instances of the object category. Additional questions may also be generated to confirm the presence of any instances that meet these relationship criteria.

Additionally, if there are no effective relationship triplets are detected, we create questions to verify the absence of specific relationships between randomly chosen pairs of object categories. These questions are designed to confirm whether certain types of objects and relationships are absent from the image. Each question is answered \textbf{negatively} to reflect the absence of such relationships.

\subsection{Complex Comprehension Tasks}

\begin{figure*}[!h]
    \centering
        \includegraphics[width=\textwidth]{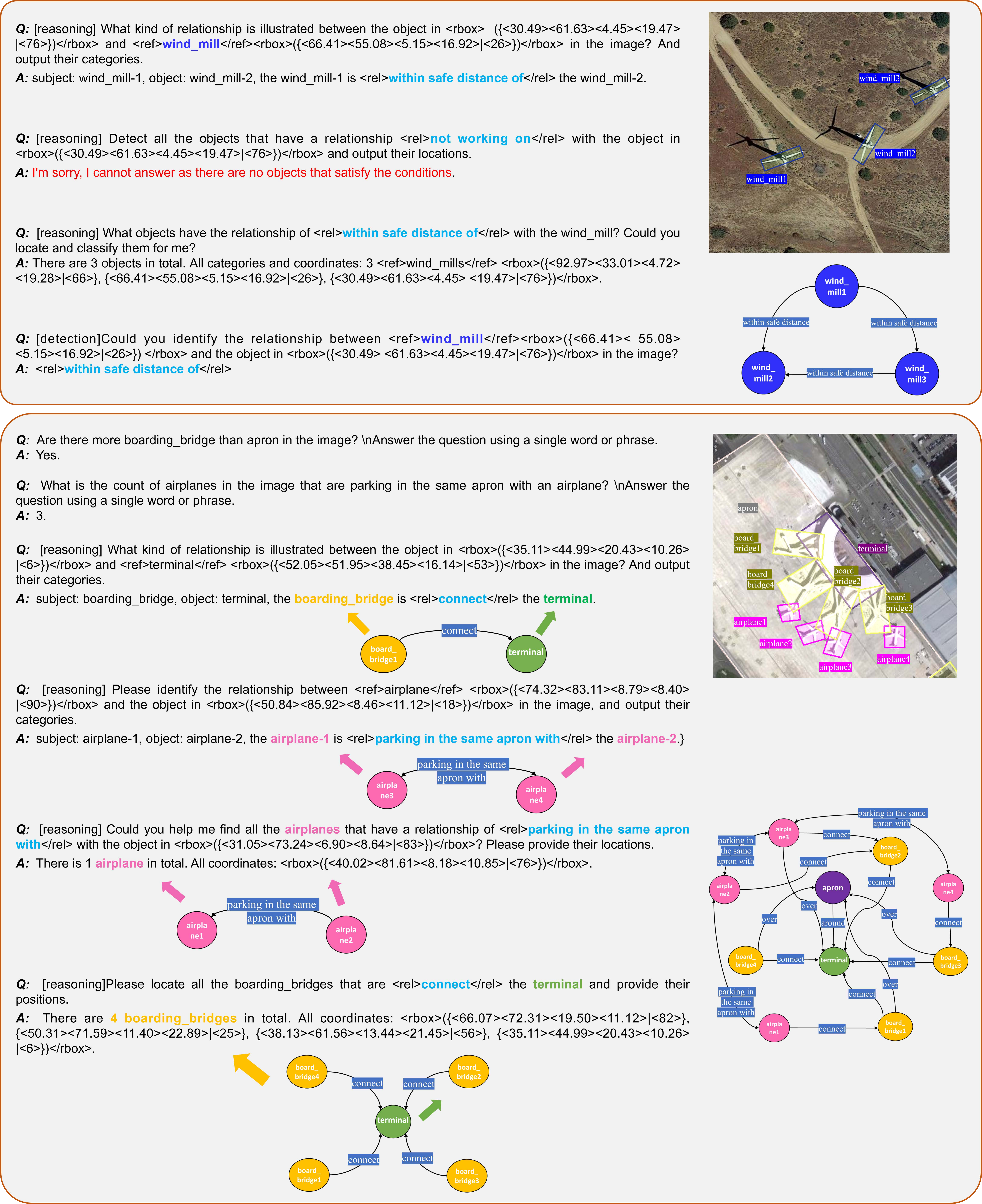}\\
\caption{An example of multi-turn conversation for a single image. It can be seen that by using user-inputted regions and relationships as aids, the model can identify the target based on the perceived scene graph.}
\label{fig:multi_vis}
\end{figure*}

In addition to the basic explanations provided in the main paper, we believe that our complex comprehension tasks proposed here surpass existing region-level tasks in several aspects. The instruction templates for constructing questions for Relation Detection, Relation Reasoning, Object Detection and Object Reasoning tasks are shown in Table~\ref{tab:complex_instruction1} and Table~\ref{tab:complex_instruction2}. For Region-Level SGG and Image-Level SGG tasks, we use the same instructions for detailed region caption and detailed image caption, respectively, but add the prefix [grounding] for them. More visual examples can be found in Figure~\ref{fig:multi_vis}.

\begin{table*}[!h]

\caption{
The templates used for constructing relation detection and relation reasoning tasks, as well as the negative answers. The content inside the special symbols in the templates will be replaced with specific target information or relationships.
}
\vspace{0.5em}
\centering
    
\begin{minipage}{\linewidth}
\begin{tcolorbox} 
\begin{tabular}{p{\linewidth}}

\VarSty{\bf Templates of Relation Detection Task:} \\
\vspace{-0.6em}
\begin{itemize}[left=0pt]
    \item What relationships exist between <category1> and <category2> in the image?
    \item Describe all existing relationships between <category1> and <category2> in the image.
    \item Describe all relationships between <category1> and <category2> in the image.
    \item Detect all existing relationships between <category1> and <category2>.
    \item Could you identify and explain the relationships between <category1> and <category2> in the image?
    \item Please analyze and describe all the relationships between <category1> and <category2> depicted in the image.
    \item Give all relationships existing between <category1> and <category2> in the image.
    \item Could you please describe all the relationships between <category1> and <category2>?
    \item In this image, can you describe all the relationships between <category1> and <category2> for me?
\end{itemize}
\vspace{1em}
\VarSty{\bf Templates of Relation Reasoning Task:} \\
\vspace{-0.6em}
\begin{itemize}[left=0pt]
    \item Please identify the relationship between <subject in> <region1> and <object in> <region2> in the image, and output their categories.
    \item Describe the relationship between <subject in> <region1> and <object in> <region2> in the image, and output their categories.
    \item What is the relationship between <subject in> <region1> and <object in> <region2> in the image? And output their categories.
    \item Could you identify the relationship between <subject in> <region1> and <object in> <region2> in the image? And output their categories.
    \item What type of relationship exists between <subject in> <region1> and <object in> <region2> in the image? And output their categories.
    \item What kind of relationship is illustrated between <subject in> <region1> and <object in> <region2> in the image? And output their categories.
    \item Could you help me understand the relationship between <subject in> <region1> and <object in> <region2> in the image? And output their categories.
\end{itemize}
\vspace{0.9em}
\VarSty{\bf Templates of Negative Answer:} \vspace{0.5em}\\
Type 1 "I'm sorry, I cannot answer as the given image does not contain any given objects." \\
Type 2 "I'm sorry, I cannot answer as the relationship is unclear in the image."\\
Type 3 "I'm sorry, I cannot answer as there are no objects that satisfy the conditions."

\end{tabular}
\end{tcolorbox}
\end{minipage}

\label{tab:complex_instruction1}
\end{table*}
\clearpage
\begin{table*}[!h]

\centering
\caption{
The templates used for constructing corpora for the object detection and object reasoning tasks. The content inside the special symbols in the templates will be replaced with specific target information or relationships.
}
\vspace{0.5em}
\begin{minipage}{\linewidth}
\begin{tcolorbox} 
\begin{tabular}{p{\linewidth}}

\VarSty{\bf Templates of Object Detection Task:} \\
\vspace{-0.5em}
\begin{itemize}[left=0pt]
    \item Can you locate all the <category> in the image?
    \item Could you help me find all the <category> in the image? Please provide their locations.
    \item Detect all the <category> in the image and output their locations.
    \item Detect all the <category> and output their locations.
    \item Provide the coordinates of all <category> in the image.
    \item Can you find and mark the positions of all the <category> in the given image?
    \item Please detect all the <category> in the image and output their locations.
    \item Locate and list the positions of all <category> that appear in the image.
    \item Identify and provide the coordinates of all <category> in the image.
    \item Identify all the <category> and mark their locations.
    \item I need you to detect and locate all <category> present in the image.
    \item Detect the locations of all <category> objects in the provided image.
    \item Please locate all the <category> in the given image.
\end{itemize}
\vspace{1em}
\VarSty{\bf Templates of Object Reasoning Task:} \\
\vspace{-0.5em}
\begin{itemize}[left=0pt]
    \item Assist me in locating and classifying all the <subject> <relation> the <object>[ in <region>].
    \item I want to know the coordinates and categories of all the <subject> <relation> the <object>[ in <region>].
    \item There are some <subject> that are <relation> the <object>[ present at <region>]. Could you tell me their locations and categories?
    \item Please locate and categorize all the <subject> that are <relation> the <object>[ in location <region>].
    \item Find all the <subject> that have a relationship of <relation> with the <object>[ present at <region>]. Can you give me their positions and categories?
    \item Your task is to locate all <subject> that have a relationship <relation> with <object>[ in location <region>] and classify them.
    \item I need you to locate and categorize all <subject> that have a relationship <relation> with <object>[ in location <region>].
    \item Could you help me find all the <subject> that have a relationship of <relation> with the <object>[ in <region>]? Please provide their locations and categories.
    \item Provide the coordinates and categories of all <subject> that are <relation> the <object>[ present at <region>] in the image.
    \item Find the <subject> that have a relationship of <relation> with the <object>[ in <region>]. Can you give me their positions and categories?
    \item What <subject> have the relationship of <relation> with the <object>[ present at <region>]? Could you locate and classify them for me?
    \item Please find and classify all <subject> that has a relationship <relation> with <object>[ in <region>].
    \item Output the positions and categories of all <subject> that have a relationship of <relation> with the <object>[ present at <region>].
    \item Please locate and classify all the <subject> that are <relation> the <object>[ present at <region>].
\end{itemize}

\end{tabular}
\end{tcolorbox}
\end{minipage}

\label{tab:complex_instruction2}
\end{table*}
\clearpage

\clearpage

\section{Details about Datasets}

\subsection{Additional Datasets for Training}
We additionally collect and convert about 365k instruction samples from existing public datasets, the specific source data quantities are shown in Table~\ref{tab:add_data_num}, and the specific conversion methods are described below. For GeoChat-Instruction~\cite{kuckreja2023geochat}, we select its image description and multi-turn conversation portions based on the DOTA~\cite{ding2021object}, DIOR~\cite{cheng2022anchor}, and FAIR1M~\cite{sun2022fair1m} datasets to enhance the overall perception of the images.

\begin{table}[!h]
\renewcommand{\arraystretch}{1.4}
\caption{Details of instruction tuning samples in additional datasets. The abbreviations and their full meanings are as follows: SC: scene classification, VQA: visual question answering, OD: object detection, IC: image caption, MT: multi-turn conversation.}
\vspace{-2pt}
\centering
\setlength{\tabcolsep}{10pt} 
\resizebox{\textwidth}{!}{ 
\begin{tabular}{cccc}
\toprule
Dataset             & Samples & Task Type  & Specific Format \\ \hline
NWPU-RESISC-45~\cite{cheng2017remote}      & 9000    & \multirow{3}{*}{SC} & \multirow{3}{*}{\makecell{Classify the given image\\ in one of the following classes.}}  \\
UCM-Landuse~\cite{qu2016deep}         & 2100    &              &                              \\
RSITMD~\cite{yuan2022exploring}              & 948     &               &                             \\\hline
EarthVQA~\cite{wang2024earthvqa}            & 145368  & \multirow{3}{*}{VQA}& \multirow{3}{*}{Answer in one word or a short phrase.} \\
RSVQA-LR~\cite{lobry2020rsvqa}            & 47173   &       &                                     \\
FloodNet-VQA\cite{rahnemoonfar2021floodnet}        & 4056    &        &            \\\hline
DOTA-v2.0~\cite{ding2021object}        & 20000    &   \multirow{2}{*}{OD}     &     \multirow{2}{*}{[detection]}       \\
FAIR1M~\cite{sun2022fair1m}        & 40000  &        &            \\ \hline
Geochat-Instruction~\cite{kuckreja2023geochat} & 96355  & IC, MT & - \\
\bottomrule
\end{tabular}
}

\label{tab:add_data_num}
\end{table}

\subsection{Datasets for Evaluation}
Details of the public datasets used for evaluation are shown in Table~\ref{tab:add_data_num_eval}. If a dataset has a clearly defined test set, we use its test set for zero-shot or fine-tuning performance evaluation. Otherwise, we typically use the entire dataset for evaluation.

\begin{table}[!h]
\caption{Details of the public datasets used for evaluation.}
\vspace{-3pt}
\renewcommand{\arraystretch}{1.4}
\centering
\setlength{\tabcolsep}{12pt} 
\resizebox{\textwidth}{!}{ 
\begin{tabular}{ccccc}
\toprule
Dataset             &  Task Type  & Image Size & Resolution (m) & Categories \\ \hline
SIRI-WHU~\cite{zhu2016bag}      &  \multirow{4}{*}{SC} & 200$\times$200 & 2 &12 \\
AID~\cite{xia2017aid}        &  & 600$\times$600 & 0.5-8 &30                       \\
WHU-RS19~\cite{dai2010satellite}              & & 600$\times$600 & 0-0.5 &19                    \\
AID-multi~\cite{hua2019relation}            &  & 600$\times$600 & 0.5-8 &17   \\ \hline
RSVQA-HR~\cite{lobry2020rsvqa}            &   \multirow{2}{*}{VQA} &  256$\times$256  &  10 &-           \\
RSVQA-LR~\cite{lobry2020rsvqa}        &  &  512$\times$512  & 0.15 &-    \\
\bottomrule
\end{tabular}
}

\label{tab:add_data_num_eval}
\end{table}
\section{Detailed Results on Benchmarks}
First, we provide additional details on the experimental setup. We use LoRA fine-tuning with the rank \( r \) set to 64. We utilize 4 GPUs for training with the total batch size set to 128 and the learning rate is \( 1 \times 10^{-6} \). The `cosine' learning rate schedule is employed with a warmup ratio of 0.03, and the training was conducted for 1 epoch.

\subsection{FIT-RSFG Benchmark}

We list the detailed results of the FIT-RSFG benchmark based on the FIT-RS dataset and the evaluation metrics used. The detailed results are shown in Table~\ref{tab:FIT_res_detail2}. The IoU threshold for evaluating the mAP metric is set to 0.25, considering that in the SGG task, the correctness of the category is more emphasized for the triplet matching. After fine-tuning, the model's performance on some challenging tasks improved significantly. Moreover, the overall accuracy distribution aligns with the progressive difficulty we designed for the tasks.

\begin{table}[!h]

\renewcommand{\arraystretch}{1.4}
\centering
\caption{Detailed result of the FIT-RSFG benchmark. For detailed image captions, the test set is selected from the descriptions generated by GPT-4. Count-Acc, Pre-Acc, Comp-Acc, and Rel-Acc represent the accuracy of four types of questions (count, presentation, comparison, and relationship).}
\setlength{\tabcolsep}{12pt} 
\resizebox{\textwidth}{!}{
\begin{tabular}{cc|c|c c c}
\toprule
\textbf{Task Type}     & \textbf{Task}     & \textbf{Metric}         & \thead{LLaVA1.5-7B\\(ZS)} & \thead{GeoChat\\(ZS)} & \thead{SkySenseGPT\\(FT)} \\ \hline
\multirow{6}{*}{IC}  & \multirow{6}{*}{Detailed Image Caption}                             & BLEU-1         & 15.38      & 8.79        &   27.31       \\
&& BLEU-2         & 13.05      & 4.36        &   15.64       \\
&& BLEU-3         & 6.23      & 1.88       &   7.95       \\
&& BLEU-4         & 2.93      & 0.83        &   4.32       \\
&& METEOR         & 10.55      & 4.23        &   12.91       \\
& &ROUGE\_L         & 20.57      & 10.55        &   23.16       \\
& &GPT-eval         & 4.30      & 4.06        &   6.00        \\ \hline
\multirow{6}{*}{RC} & \multirow{6}{*}{Detailed Region Caption}          
& BLEU-1         & 14.70      & 14.55       &   75.82       \\
&& BLEU-2         & 6.63      & 8.66       &   71.42       \\
&& BLEU-3         & 2.33      & 4.63       &   67.71       \\
&& BLEU-4         & 0.63      & 1.97       &   64.05       \\
&& METEOR          & 10.66      & 8.80       &   43.69       \\
&& ROUGE\_L         & 15.67      & 14.02       &   71.86       \\\hline
\multirow{5}{*}{VQA}                        & \multirow{5}{*}{\thead{Visual Question \\Answering}}    & Count-Acc     & 27.75      & 31.64       &     46.49     \\
     &   & Pre-Acc     & 81.91      & 51.70       &   92.82       \\
     &  & Comp-Acc    & 55.29      & 64.13       &  82.54        \\
     &  & Rel-Acc     & 69.40       & 66.40       &   90.18       \\
 &    & Ave Acc        & 58.59      & 53.47       &  79.76        \\\hline
\multirow{4}{*}{SC} & \multirow{4}{*}{\thead{Multi-Label\\Scene Classification}} & Scene Acc & 56.18      & 42.27       &    82.23      \\
                                    &        & Obj-Precision  & 19.04      & 9.10        &  77.28        \\
                                    &        & Obj-Recall     & 9.69       & 41.89       &   66.40       \\
               &                             & Obj-F1-Score   & 12.84      & 14.95       &  71.43       \\\hline

\multirow{6}{*}{CC} & Relation Detection     & F1-Score       & Failed     & Failed      &   88.68       \\

   &  Relation   Reasoning                          & F1-Score       & Failed     & Failed      &   74.33       \\ 
&Object   Detection                          & mAP            & Failed     & 6.13        &    27.40      \\
&Object   Reasoning                         & mAP            & Failed     & Failed      &    5.71      \\
& \multirow{2}{*}{Region-Level SGG}       & Recall         & Failed     & Failed      &  17.01        \\
                                      &      & Mean Recall    & Failed     & Failed      &   4.05       \\
&\multirow{3}{*}{Image-Level SGG}             & Recall         & Failed     & Failed      &   9.60       \\
                                     &       & Mean Recall    & Failed     & Failed      & 3.59         \\
                                  &          & mAP            & Failed     & Failed      & 12.99 \\\hline       

MT & Multi-Turn Conv  & GPT-eval & 2.74     & 3.64      &   6.60       \\
\bottomrule
\end{tabular}
}
\vspace{5pt}

\label{tab:FIT_res_detail2}
\end{table}

\subsection{FIT-RSRC Benchmark}

\begin{figure*}[!h]
    \centering
        \includegraphics[width=\textwidth]{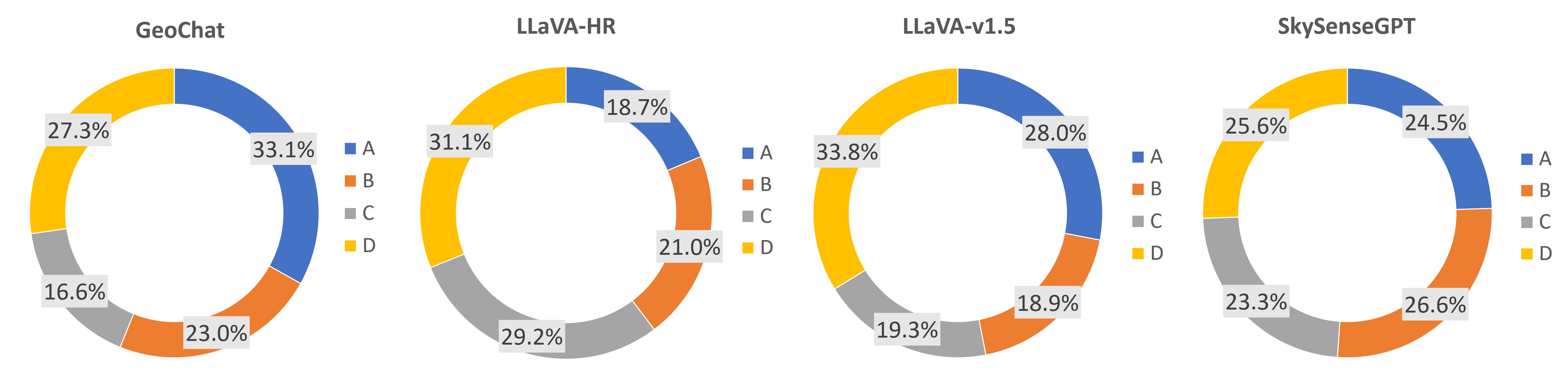}\\
\caption{The choice distribution of predictions of different LMMs, all using CirularEval records.}
\label{fig:option_distribution}
\end{figure*}

\vspace{-3pt}
In the FIT-RSRC benchmark, we construct four types of questions (relationship, object, subject, existence) each accounting for 25\%, with negative answers distributed within each type. The length of each question option is 4-5, with 87.5\% of questions having 4 options and 12.5\% having 5 options. To better analyze the prediction accuracy of the evaluated LMMs, we count the distribution of answers for all 4-option questions across the four models, as shown in Figure~\ref{fig:option_distribution}. It can be seen that the existing LMMs show a bias towards a certain option (option D), whereas SkySenseGPT exhibits less bias.

Additionally, we present the carefully crafted preset commonsense word lists, created by human experts with the assistance of GPT-4. These lists include the subject-relation and relation-object word lists, as shown in Table~\ref{tab:common_word_list} and Table~\ref{tab:common_word_list2}. When generating high-quality distractor options, all generated content is cross-checked against these word lists to ensure that incorrect options cannot be identified solely based on the text. This ensures the high quality of the FIT-RSRC benchmark.

\begin{table*}[!h]

\centering
\caption{
The subject-relation word list created by experts with the assistance of GPT-4. All distractor options in FIT-RSRC need to exist in this list to ensure that the LMM cannot deduce the answer solely from the text.
}
\begin{minipage}{\linewidth}
\begin{tcolorbox} 
\begin{tabular}{p{\linewidth}}
\VarSty{\bf Commonsense Subject-Relation List:} \\
\vspace{-0.5em}
{\bf storehouse:} adjacent, connect, over, co-storage with, around, not co-storage with \\
{\bf breakwater:} connect, over, adjacent, through, converge, intersect, around \\
{\bf dock:} connect, over, adjacent, through, converge, intersect, around \\
{\bf airplane:} connect, over, within safe distance of, parallelly parked on, isolatedly parked on, randomly parked on, run along, not run along, parking in the same apron with, parking in the different apron with, parked alongside with, not parked alongside with, running along the different taxiway with, running along the same taxiway with, approach, away from, within danger distance of, pass under, pass through, pass across \\
{\bf boarding bridge:} connect, over, adjacent, through, converge, intersect, around \\
{\bf runway:} connect, over, adjacent, through, converge, intersect, around \\
{\bf taxiway:} connect, over, adjacent, through, converge, intersect, around \\
{\bf terminal:} connect, over, adjacent, through, converge, intersect, around \\
{\bf apron:} connect, adjacent, through, converge, intersect, around \\
{\bf truck:} within safe distance of, driving in the same direction with, parallelly parked on, isolatedly parked on, parked alongside with, not parked alongside with, approach, away from, within danger distance of, incorrectly parked on, in the same parking with, in the different parking with, driving in the opposite direction with, driving in the different lane with, driving in the same lane with, driving alongside with, drive toward, pass across, drive off \\
{\bf car:} within safe distance of, driving in the same direction with, parallelly parked on, isolatedly parked on, parked alongside with, not parked alongside with, approach, away from, within danger distance of, incorrectly parked on, in the same parking with, in the different parking with, driving in the opposite direction with, driving in the different lane with, driving in the same lane with, driving alongside with, drive toward, pass across, drive off \\
{\bf cooling tower:} violently emit, slightly emit, exhaust to, supply to \\
{\bf chimney:} violently emit, slightly emit \\
{\bf vapor:} connect \\
{\bf smoke:} connect \\
{\bf genset:} over, exhaust to, supply to, directly transmit electricity to \\
{\bf coal yard:} adjacent, supply to, over, through, intersect \\
{\bf lattice tower:} within different line of, within same line of, directly connected to, indirectly connected to, directly transmit electricity to, indirectly transmit electricity to \\
{\bf substation:} within same line of, directly connected to, directly transmit electricity to, indirectly transmit electricity to \\
{\bf wind mill:} within safe distance of \\
{\bf flood dam:} adjacent, over, connect, intersect \\
{\bf ground track field:} around, over, through, converge, intersect \\
{\bf basketball court:} around, over, through, converge, intersect \\
{\bf engineering vehicle:} working on, not working on \\
{\bf soccer ball field:} around, over, through, converge, intersect \\
{\bf tennis court:} around, over, through, converge, intersect \\
{\bf tower crane:} working on, not working on \\
{\bf containment vessel:} co-storage with, not co-storage with, supply to \\

\end{tabular}
\end{tcolorbox}
\end{minipage}
\label{tab:common_word_list}
\end{table*}

\clearpage

\begin{table*}[!h]

\centering
\caption{
Part of the relation-object word list created by experts with the assistance of GPT-4. All distractor options in FIT-RSRC need to exist in this list to ensure that the LMM cannot deduce the answer solely from the text.
}
\begin{minipage}{\linewidth}
\begin{tcolorbox} 
\begin{tabular}{p{\linewidth}}
\VarSty{\bf Commonsense Relation-Object List:} \\
\vspace{-0.5em}
{\bf parallelly docked at:} dock, breakwater \\
{\bf isolatedly docked at:} dock, breakwater \\
{\bf connect:} tank, storehouse, breakwater, dock, boarding bridge, runway, taxiway, terminal, apron, vapor, smoke, substation, toll gate, bridge, intersection, roundabout, genset, cooling tower, chimney, lattice tower, wind mill, containment vessel \\
{\bf over:} ship, boat, crane, goods yard, breakwater, dock, runway, taxiway, terminal, apron, truck parking, car parking, bridge, genset, coal yard, cement concrete pavement, flood dam, gravity dam, ground track field, basketball court, foundation pit, intersection, soccer ball field, tennis court, arch dam, roundabout, baseball diamond, stadium \\
{\bf co-storage with:} containment vessel, tank, storehouse, goods yard, coal yard, substation, genset, cooling tower, chimney \\
{\bf within safe distance of:} ship, boat, airplane, truck, car, wind mill \\
{\bf randomly docked at:} dock, breakwater \\
{\bf docking at the same dock with:} ship, boat \\
{\bf docked alongside with:} ship, boat \\
{\bf docking at the different dock with:} ship, boat \\
{\bf driving in the same direction with:} ship, boat, engineering vehicle, truck, car \\
{\bf parallelly parked on:} runway, taxiway, apron, truck parking, car parking \\
{\bf isolatedly parked on:} runway, taxiway, apron, truck parking, car parking \\
{\bf randomly parked on:} runway, taxiway, apron, truck parking, car parking \\
{\bf run along:} runway, taxiway \\
{\bf adjacent:} flood dam, crane, goods yard, storehouse, breakwater, dock, runway, taxiway, terminal, apron, genset, coal yard \\
{\bf through:} runway, taxiway {\bf converge:} runway, taxiway  
{\bf intersect:} runway, taxiway \\
{\bf not run along:} runway, taxiway \\
{\bf parking in the same apron with:} airplane \\
{\bf parking in the different apron with:} airplane \\
{\bf parked alongside with:} engineering vehicle, crane, airplane, truck, car \\
{\bf not parked alongside with:} engineering vehicle, crane, truck, airplane, car \\
{\bf running along the different taxiway with:} airplane, engineering vehicle, crane, truck, car \\
{\bf around:} ground track field, basketball court, tank, soccer ball field, taxiway, terminal, car parking, truck parking \\
{\bf not co-storage with:} containment vessel, tank, storehouse \\
{\bf running along the same taxiway with:} airplane, engineering vehicle, crane, truck, car \\
{\bf approach:} intersection, dock, breakwater, runway, taxiway, roundabout, apron, gas station, bridge, cement concrete pavement, toll gate \\
{\bf away from:} ship lock, intersection, dock, breakwater, runway, taxiway, roundabout, apron, gas station, truck parking, car parking, bridge, cement concrete pavement, toll gate \\
{\bf within danger distance of:} ship, boat, airplane, truck, car \\
{\bf running along the different runway with:} airplane, engineering vehicle, crane, truck, car \\
{\bf incorrectly parked on:} truck parking, car parking \\
{\bf in the same parking with:} truck, car, engineering vehicle, crane, airplane \\
{\bf in the different parking with:} truck, car, engineering vehicle, crane, airplane \\
{\bf not docked alongside with:} ship, boat \\
{\bf driving in the opposite direction with:} ship, boat, truck, car, engineering vehicle, crane, airplane \\
{\bf driving in the different lane with:} truck, car, engineering vehicle, crane, airplane \\
{\bf driving in the same lane with:} truck, car, engineering vehicle, crane, airplane \\
{\bf docking at the same breakwater with:} ship, boat \\
{\bf driving alongside with:} truck, car, engineering vehicle, crane

\end{tabular}
\end{tcolorbox}
\end{minipage}
\label{tab:common_word_list2}
\end{table*}

\clearpage

\end{document}